\pdfoutput=1

\documentclass[11pt]{article}

\usepackage[final]{coling}

\usepackage{times}
\usepackage{latexsym}

\usepackage[T1]{fontenc}

\usepackage[utf8]{inputenc}

\usepackage{microtype}

\usepackage{hyperref}
\usepackage{inconsolata}
\usepackage{tcolorbox}
\tcbuselibrary{skins, breakable, theorems}
\usepackage{booktabs} 
\usepackage{arydshln}
\usepackage{makecell}
\usepackage{colortbl}
\usepackage{tablefootnote}
\usepackage{multirow}
\usepackage{subcaption}
\usepackage{graphicx}
\usepackage{float}
\usepackage{xcolor}
\usepackage{amsmath}
\usepackage{amsfonts}
\usepackage{enumitem}

\usepackage{siunitx}
\usepackage{bm}

\title{ACE-$M^3$: Automatic Capability Evaluator for Multimodal Medical Models}

\author{
 \textbf{Xiechi Zhang\textsuperscript{1}},
 \textbf{Shunfan Zheng\textsuperscript{1}},
 \textbf{Linlin Wang\textsuperscript{1}\thanks{Corresponding author.}},
 \textbf{Gerard de Melo\textsuperscript{2}},
\\
 \textbf{Zhu Cao\textsuperscript{3}},
 \textbf{Xiaoling Wang\textsuperscript{1}},
 \textbf{Liang He\textsuperscript{1}}
\\
\\
 \textsuperscript{1}East China Normal University\\
 \textsuperscript{2}Hasso Plattner Institute/University of Potsdam
\\
 \textsuperscript{3}East China University of Science and Technology
\\
 \small{
   \textbf{Correspondence:} \href{mailto:llwang@cs.ecnu.edu.cn}{llwang@cs.ecnu.edu.cn}
 }
}

\begin{document}
\maketitle
\begin{abstract}
As multimodal large language models (MLLMs) gain prominence in the medical field, the need for precise evaluation methods to assess their effectiveness has become critical. While benchmarks provide a reliable means to evaluate the capabilities of MLLMs, traditional metrics like ROUGE and BLEU employed for open domain evaluation only focus on token overlap and may not align with human judgment. Although human evaluation is more reliable, it is labor-intensive, costly, and not scalable. LLM-based evaluation methods have proven promising, but to date, there is still an urgent need for open-source multimodal LLM-based evaluators in the medical field. To address this issue, we introduce ACE-$M^3$, an open-sourced \textbf{A}utomatic \textbf{C}apability \textbf{E}valuator  for \textbf{M}ultimodal \textbf{M}edical \textbf{M}odels specifically designed to assess the question answering abilities of medical MLLMs. It first utilizes a branch-merge architecture to provide both detailed analysis and a concise final score based on standard medical evaluation criteria. Subsequently, a reward token-based direct preference optimization (RTDPO) strategy is incorporated  to save training time without compromising performance of our model. Extensive experiments have demonstrated the effectiveness of our ACE-$M^3$ model\footnote{\url{https://huggingface.co/collections/AIUSRTMP/ace-m3-67593297ff391b93e3e5d068}} in evaluating the capabilities of medical MLLMs. 
\end{abstract}

\section{Introduction}
The emergence of increasingly powerful large language models (LLMs) has driven significant advances in Multimodal LLMs~(MLLMs; \citealp{liu2024visual, liu2024llavanext}), particularly in specialized domains such as the medical field~\cite{li2024llava, yang2024advancing, pellegrini2023radialog, moor2023med}. This progress has underscored the urgent need for reliable evaluation systems to assess and compare their performance. However, comprehensively evaluating the capabilities of various medical MLLMs remains a formidable challenge due to the necessity of medical expert knowledge and the substantial workload involved~\cite{singhal2023large, chang2024survey, yin2024survey}.


Although benchmarks like Path-VQA~\cite{he2020pathvqa} can be used to assess the capabilities of medical MLLMs~\cite{pmlr-v225-moor23a,li2024llava}, they still use traditional metrics such as ROUGE~\cite{2004ROUGE} and BERTScore~\cite{zhangbertscore} to perform open-ended generation evaluations, which may fail to align with humans, since they predominantly consider lexical or semantic matches. Human evaluation is often used to gauge the efficacy of medical MLLMs, but this approach is labor-intensive, time-consuming, and impractical for large-scale applications, especially when medical expertise is needed~\cite{xu2023medgpteval}. 

Leveraging LLMs as evaluators represents an innovative and promising approach~\cite{zheng2024judging}. However, 
although proprietary models like GPT-4 can provide detailed assessments~\cite{nori2023capabilities}, they are hampered by a lack of transparency and reproducibility as well as potential privacy leakage. Moreover, the inability to correct biases~\cite{zack2024assessing} hinders the application of closed-source MLLMs. Meanwhile, existing open-source evaluators are all text-only evaluation models and are designed for general domain assessment purposes~\cite{wang2024pandalm, ligenerative}, which highlights an urgent need for an open-sourced multimodal evaluation model that can provide detailed and reliable analysis with corresponding scores.

To address this urgent need, we propose a multimodal medical evaluation model named ACE-$M^3$, which can provide detailed and reliable evaluations. Specifically, we first build an instruction dataset based on existing benchmarks with reliable evaluation criteria using powerful LLMs. Subsequently, we employ a branch-merge architecture with an Efficient Reward Token-based Direct Preference Optimization (Efficient-RTDPO) training strategy to build the ACE-$M^3$ model based on the MedLlama2\footnote{https://huggingface.co/llSourcell/medllama2\_7b} model using the collected instruction dataset. The branch-merge architecture enables ACE-$M^3$ to provide detailed analysis and a concise score for each response, while the Efficient-RTDPO strategy saves training time without compromising evaluation accuracy.

In summary, our contributions can be articulated as follows:
\begin{itemize}[noitemsep,nolistsep]
\item By leveraging reliable evaluation criteria and powerful LLMs, we meticulously curate a multimodal medical instruction dataset, which can facilitate the development of multimodal medical evaluation models.
\item  To the best of our knowledge, we are the first to propose a multimodal medical evaluation model. The proposed model, ACE-$M^3$, can provide detailed analysis and concise scores of medical MLLMs by using a branch-merge architecture. An Efficient-RTDPO training strategy is proposed to save training time without harming the evaluation accuracy.
\item  Extensive experiments and discussions have been performed to prove the effectiveness of the ACE-$M^3$ model and the Efficient-RTDPO training strategy.
\end{itemize}

\section{Methodology}
In this section, we first delineate the evaluation criteria proposed for model training and dataset construction. Subsequently, we introduce the branch-merge evaluation framework and implementation details of the models, complete with a training strategy named Efficient-RTDPO, which saves training time without impeding the evaluation accuracy. Additionally, we detail the methodologies employed in constructing the instruction datasets, elaborating on the selection of QA datasets and the mechanisms utilized to secure reliable evaluations.

\begin{figure*}[htbp]
    \centering
    \hspace*{7mm}\includegraphics[width=0.9\textwidth]{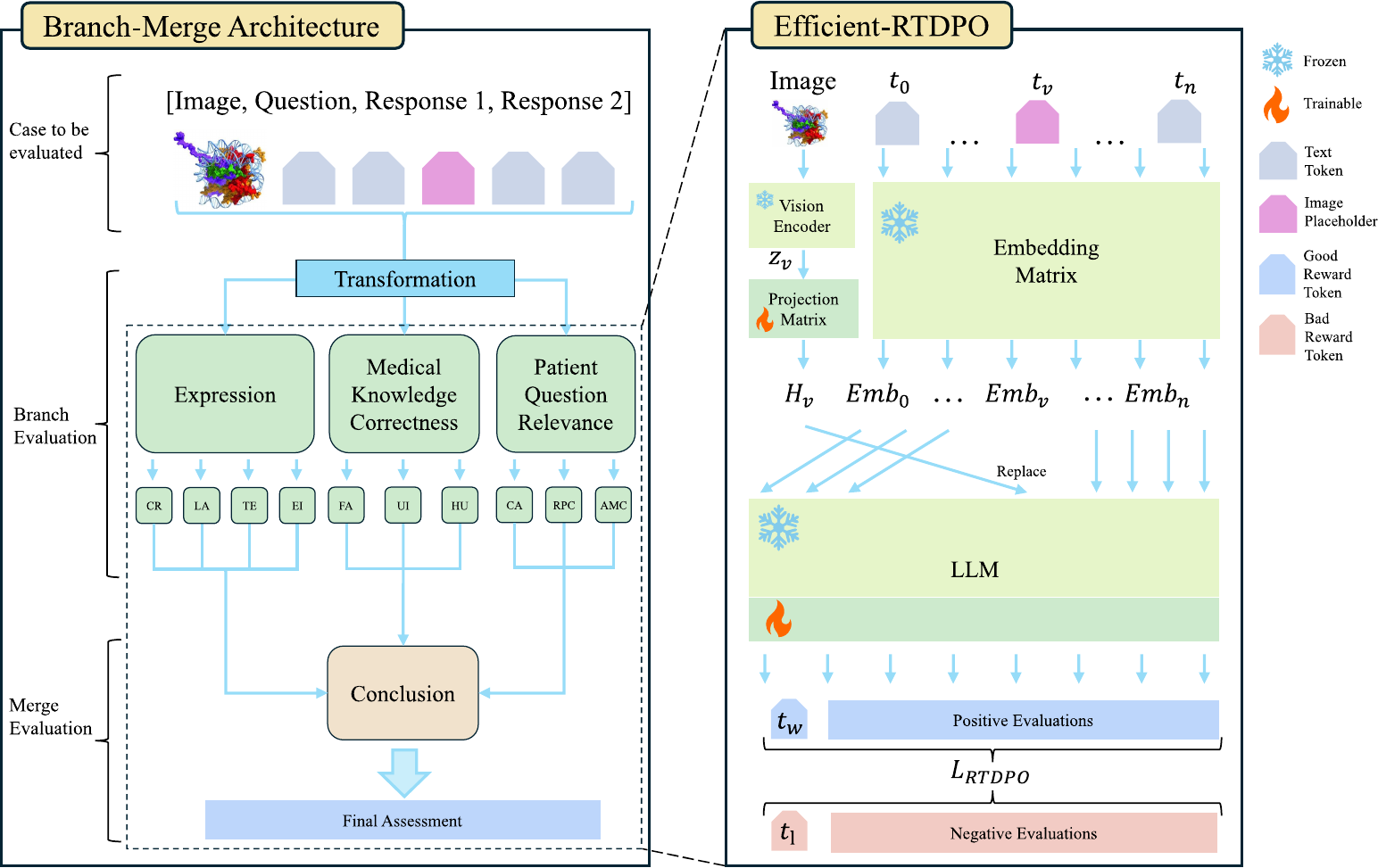}
    \caption{Framework and training details of our multimodal evaluation model ACE-$M^3$.}
    \label{fig:framework}
    \vspace{-3mm}
\end{figure*}

\subsection{Reliable Evaluation Criteria}
\label{Reliable Evaluation Criteria}
Reliable evaluation criteria are crucial to evaluate the performance of LLMs. We therefore invited three professional annotators to discuss and meticulously formulate detailed and reliable evaluation criteria concerning the following aspects (more detailed explanations of each criterion are given in Appendix~\ref{Detailed Criteria}):
\begin{itemize}
    \item \textbf{Expression (EXP)}: $(1)$ Clarity of Response (CR), $(2)$ Language Appropriateness (LA), $(3)$ Tone and Empathy (TE), and $(4)$ Expression Integrity (EI).
    \item \textbf{Medical Knowledge Correctness (MKC)}: $(1)$ Factual Accuracy (FA), $(2)$ Up-to-date Information (UI), and $(3)$ Handling Uncertainty (HU).
    \item \textbf{Patient Question Relevance (PQR)}: $(1)$ Context Awareness (CA), $(2)$ Relevance to Patient’s Condition (RPC), and (3) Addressing Multiple Concerns (AMC).
\end{itemize}

\subsection{Model Overview and Details}
In this subsection, we first elucidate the architecture of our model, subsequently delving into the implementation details, including the methodologies for processing image inputs and the details of our Efficient-RTDPO training strategy.

\subsubsection{Branch-Merge Architecture}
As shown in the left part of Figure~\ref{fig:framework}, we employ a branch-merge architecture that consists of three sub-domain evaluation models and a conclusion evaluation model. For every instance to be evaluated, we compute an overall conclusion evaluation $E_{c}$ as follows:
\begin{align}
    E_{c} = M(v, q, r_1, r_2)
\end{align}
\noindent where $M$ represents the model, $v$ and $q$ stand for the image input and question, and $r_1$, $r_2$ refer to two responses of medical MLLMs. 
To achieve this, we first employ three branch-specific evaluation models $M_{s_i}$ to evaluate the instance according to the criteria in Section~\ref{Reliable Evaluation Criteria}. Subsequently, we feed the branch-specific evaluations $E_{s_i}$ and the original inputs into our conclusion model $M_c$, obtaining the final assessment $E_c$:
\begin{align}
    &E_{s_{i}} = M_{s_{i}}(v, q,r_1,r_2)\\
    &E_{c} = M_{c}(v, q,r_1,r_2,E_{s_{1}},E_{s_{2}},E_{s_{3}})
\end{align}
The prompt templates utilized in our experiments are given in Appendix~\ref{Prompt for Instruction Tuning}.

\subsubsection{Process of Image Inputs}
As depicted in the right part of Figure~\ref{fig:framework}, we adopt a method that uses a projection matrix to link the visual encoder and the large language model~\cite{liu2024visual}. Specifically, for an input image $X_v$, we first exploit a pre-trained vision encoder $M_\text{vision}$ (CLIP; \citealp{radford2021learning}) to capture the visual features of the image as $Z_v$. Subsequently, a projection weight matrix $W$ is employed to project the visual feature representation $Z_v$ into the hidden state $H_v$, which has  the same dimensionality as the embedding space of the large language model:
\begin{align}
    &Z_v = M_\text{vision}(X_v) \\
    &H_v = W_\text{proj}(Z_v)
\end{align}
\noindent Then the image placeholder embedding $Emb_{v}$ is replaced by the image hidden state $H_v$. Subsequently, the image hidden state $H_v$ is concatenated with the text token embedding for evaluation generation via an LLM $M_\text{LLM}$:
\begin{align}
    E = M_\text{LLM}(\text{Emb}_{0,v-1};H_v;\text{Emb}_{v+1,n})
\end{align}
\noindent Note that the vision encoder is frozen and only the projection matrix is trained to learn how to match the LLM's representation space during training.

\begin{table*}[htbp]
  \centering
  \resizebox{\linewidth}{!}{
    \begin{tabular}{lrrrrrrrrrrrr}
    \Xhline{1.0pt}
    \multirow{2}[1]{*}{\textbf{Dataset}} & \multicolumn{3}{c}{Conclusion} & \multicolumn{3}{c}{EXP} & \multicolumn{3}{c}{MKC} & \multicolumn{3}{c}{PQR} \\
    \cmidrule(lr){2-4} \cmidrule(lr){5-7} \cmidrule(lr){8-10} \cmidrule(lr){11-13}
    \multicolumn{1}{l}{} & \multicolumn{1}{c}{Total} & \multicolumn{1}{c}{F.Q.} & \multicolumn{1}{c}{E.Q.} & \multicolumn{1}{c}{Total} & \multicolumn{1}{c}{F.Q.} & \multicolumn{1}{c}{E.Q.} & \multicolumn{1}{c}{Total} & \multicolumn{1}{c}{F.Q.} & \multicolumn{1}{c}{E.Q.} & \multicolumn{1}{c}{Total} & \multicolumn{1}{c}{F.Q.} & \multicolumn{1}{c}{E.Q.} \\
    \hline
    MedDialogue-EN & 18,022  & 17,583  & 94\%  & 18,410  & 18,295  & 94\%  & 18,410  & 18,380  & 90\%  & 18,410  & 18,144  & 90\% \\
    MedText  & 42,027  & 41,504  & 93\%  & 42,360  & 42,271  & 93\%  & 42,360  & 42,340  & 93\%  & 42,360  & 42,130  & 93\% \\
    MedBench & 1,733  & 1,701  & 89\%  & 1,837  & 1,769  & 93\%  & 1,837  & 1,774  & 91\%  & 1,837  & 1,762  & 92\% \\
    MeadowWikidoc & 8,732  & 8,258  & 94\%  & 9,569  & 8,732  & 90\%  & 9,569  & 8,736  & 93\%  & 9,569  & 8,741  & 90\% \\
    Path-VQA  & 12,077  & 12,043  & 90\%  & 13,438  & 13,101  & 95\%  & 13,438  & 13,179  & 92\%  & 13,438  & 12,610  & 94\% \\
    SLAKE & 12,285  & 12,205  & 92\%  & 14,066  & 13,515  & 91\%  & 14,066  & 14,009  & 90\%  & 14,066  & 12,818  & 91\% \\
    VQA-RAD & 3,930  & 3,918  & 90\%  & 4,488  & 4,375  & 91\%  & 4,488  & 4,463  & 90\%  & 4,488  & 4,052  & 92\% \\
    \Xhline{1.0pt}
    \end{tabular}
    }
    \caption{Detailed number of samples in each dataset's four aspects. Format Qualified (F.Q.) represents the number of samples from which the evaluation content generated by GPT-3.5-Turbo obeys the format requirements in the instruction while Evaluation Qualified (E.Q.) denotes the percentage of accurate samples in the sample inspection.}
    \label{tab:statistics}
    \vspace{-3mm}
\end{table*}

\subsubsection{Efficient-RTDPO Training Strategy}
Due to the substantial model training effort necessitated by the branch-merge framework, previous evaluation paradigms are often restricted to the form of either a single model only~\cite{wang2024pandalm, ligenerative} or inference only~\cite{saha2024branch}. We thus propose an Efficient-RTDPO training strategy that can help save training time without harming the evaluation accuracy when training a group of evaluation models. 

Specifically, we initially freeze the lower-layer parameters of the model to curtail training time, albeit at the expense of diminished performance. Then for counterbalancing the decrease in accuracy, we propose a Reward Token-based Direct Preference Optimization (RTDPO) strategy based on the commonly-used DPO strategy~\cite{rafailov2024direct}, which steers our model towards a more accurate evaluation. Specifically, we prepend the positive reward token $\bm{t_w}$ (e.g., \texttt{[Good]}) to the positive evaluations $e_w$, while the negative reward token $\bm{t_l}$ (e.g., \texttt{[Bad]}) is prepended to the negative evaluations $e_l$. We define a novel loss function as follows:
\begin{equation}
\begin{aligned}
\mathcal{L}_{\text{RTDPO}}(\pi_\theta; \pi_{\text{ref}}) = -\mathbb{E}_{(x,t_w, e_w, t_l, e_l) \sim \mathcal{D}} \Bigg[ \log \sigma 
    \Big( \beta &\\ \log \frac{\pi_\theta(\bm{t_w}, e_w \mid x)}{\pi_{\text{ref}}(\bm{t_w}, e_w \mid x)}  - \beta \log \frac{\pi_\theta(\bm{t_l}, e_l \mid x)}{\pi_{\text{ref}}(\bm{t_l}, e_l \mid x)} \Big) \Bigg] &
\end{aligned}
\end{equation}
\noindent where $x$ denotes the case to be evaluated, $\pi_{\theta}$, $\pi_{ref}$ represent the policy and reference model, respectively, and $\beta$ is a hyperparameter that controls the divergence.

Considering that the key goal of our model is to provide high-precision evaluation scores for comparison, we construct negative evaluation samples with the following methods: $(1)$ \textbf{score swapping}, i.e., swapping the scores concerning two responses for each criterion, $(2)$ \textbf{score addition}, i.e., adding two points to each criterion for both responses, and $(3)$ \textbf{score subtraction}, i.e., subtracting two points to each criterion for both responses.

\subsection{Instruction Dataset Construction}
In this subsection, we first introduce the benchmarks utilized, followed by the process to
obtain the responses $r$ and the evaluations $E$.

\subsubsection{Medical QA \& VQA Benchmarks}
To construct a multimodal instruction dataset for training the evaluation model, we employ three widely recognized visual question-answering benchmarks, Path-VQA~\cite{he2020pathvqa}, SLAKE~\cite{9434010}, and VQA-RAD~\cite{lau2018dataset}, to serve as the foundational data sources. Additionally, we integrate four commonly-used text-only datasets, MedDialogue-EN~\cite{zeng2020meddialog}, MedText\footnote{https://huggingface.co/datasets/BI55/MedText}, MedBench~\cite{cai2024medbench}, and MedicalMeadowWikidoc\footnote{https://huggingface.co/datasets/medalpaca/medical\_m-\\eadow\_wikidoc} during the construction of instructions and training phases of the text-only evaluation model. Moreover, we selectively retain only test sets in subsequent processing for datasets with excessive data to ensure the efficiency of training the evaluation model. Table~\ref{tab:dataset details} provides statistical details of the source data.

\begin{table}[h]
    \centering
    \resizebox{0.9\linewidth}{!}{
    \begin{tabular}{lrrrr}
    \Xhline{1.0pt}
    \textbf{Dataset} & \textbf{Images} & \textbf{QA Pairs} & \textbf{Open} & \textbf{Closed} \\ 
    \rowcolor{gray!25}\multicolumn{5}{l}{\textit{QA Benchmarks}} \\
    MedDialogue-EN & / & 615  & 615 & 0\\
    MedText & / & 1,412 & 1,412 & 0\\ 
    MedBench & / & 1,737 & 1,737 & 0\\ 
    MeadowWikidoc & / & 10,000 & 10,000 & 0\\ 
    \rowcolor{gray!25}\multicolumn{5}{l}{\textit{VQA Benchmarks}} \\
    Path-VQA & 858 & 6,719 & 3,357 & 3,362\\ 
    SLAKE & 642 & 7,033 & 4,252 & 2,781\\
    VQA-RAD & 516 & 2,244 & 1,053 & 1,191\\
    \Xhline{1.0pt}
    \end{tabular}
    }
    \caption{Statistics of the datasets utilized, where open means the answer to the question is open-ended while closed means the answer is yes/no.}
    \label{tab:dataset details}
    \vspace{-3mm}
\end{table}

\subsubsection{Response Generation}
We incorporate four unimodal and two multimodal medical models to generate responses for the collected questions. The specific parameters of the models used in the collection phase are detailed in Table~\ref{tab:model details}. Furthermore, a standardized query template (outlined in Appendix~\ref{Prompt for Response Collection}) and greedy decoding are systematically employed to maintain fairness across the same type 
of models while generating responses.

\begin{table}[h]
\small
\centering
\resizebox{0.7\linewidth}{!}{
\begin{tabular}{lr}
\Xhline{1.0pt}
\textbf{Model Name} & \textbf{Model Size}\\ 
\rowcolor{gray!25}\multicolumn{2}{l}{\textit{Text-only Medical LLMs}} \\
ChatDoctor~\cite{li2023chatdoctor} & 7B \\ 
MedAlpaca~\cite{han2023medalpaca} & 7B \\
MedLlama2 & 7B \\
Baize v2~\cite{xu2023baize} & 13B \\
\rowcolor{gray!25}\multicolumn{2}{l}{\textit{Image--Text Medical MLLMs}} \\
LLaVA-Med~\cite{li2024llava} & 7B \\
CheXagent~\cite{chen2024chexagent} & 8B \\
\Xhline{1.0pt}
\end{tabular}
}
\caption{Statistics of the Medical LLMs and MLLMs employed.}
\label{tab:model details}
\vspace{-4mm}
\end{table}

\subsubsection{Evaluation Collection}
Following the previous work~\cite{wang2024pandalm, chanchateval}, we adopt the common practice of distilling evaluations from powerful LLMs. Specifically, as shown in Appendix~\ref{Prompt for Evaluation Collection}, the prompt is comprised of the evaluation criterion, output format requirements, and the task definition, designed to steer GPT-3.5-Turbo (gpt-3.5-turbo-0125) to conduct evaluations. Additionally, we supply the reference answers to ensure more reliable and accurate evaluations, which also solves the issue that  GPT-3.5-Turbo cannot process image inputs. 

After collecting the evaluations, we conduct the following two-step check  to ensure data reliability. $(1)$ Format Check: we examine whether the output from GPT-3.5-Turbo adheres to the format requirements specified in the prompt and verify that the scores within the evaluation content can be successfully extracted. $(2)$ Content Verification: given the impracticality of manually checking all samples, we perform a sampling check on the evaluation content. Specifically, we invited the aforementioned annotators to form a committee and manually annotate 100 evaluation entries from each of the four evaluation aspects of each dataset. The statistics of the evaluations collected are given in Table~\ref{tab:statistics} and Figure~\ref{fig:conclusion co-occurrence}. An example of a collected evaluation can be seen in Appendix~\ref{Example of Collected Evaluation} and annotation details are provided in Appendix~\ref{Annotation Details}.

\begin{figure}[htbp]
    \centering
    \includegraphics[width=0.85\linewidth]{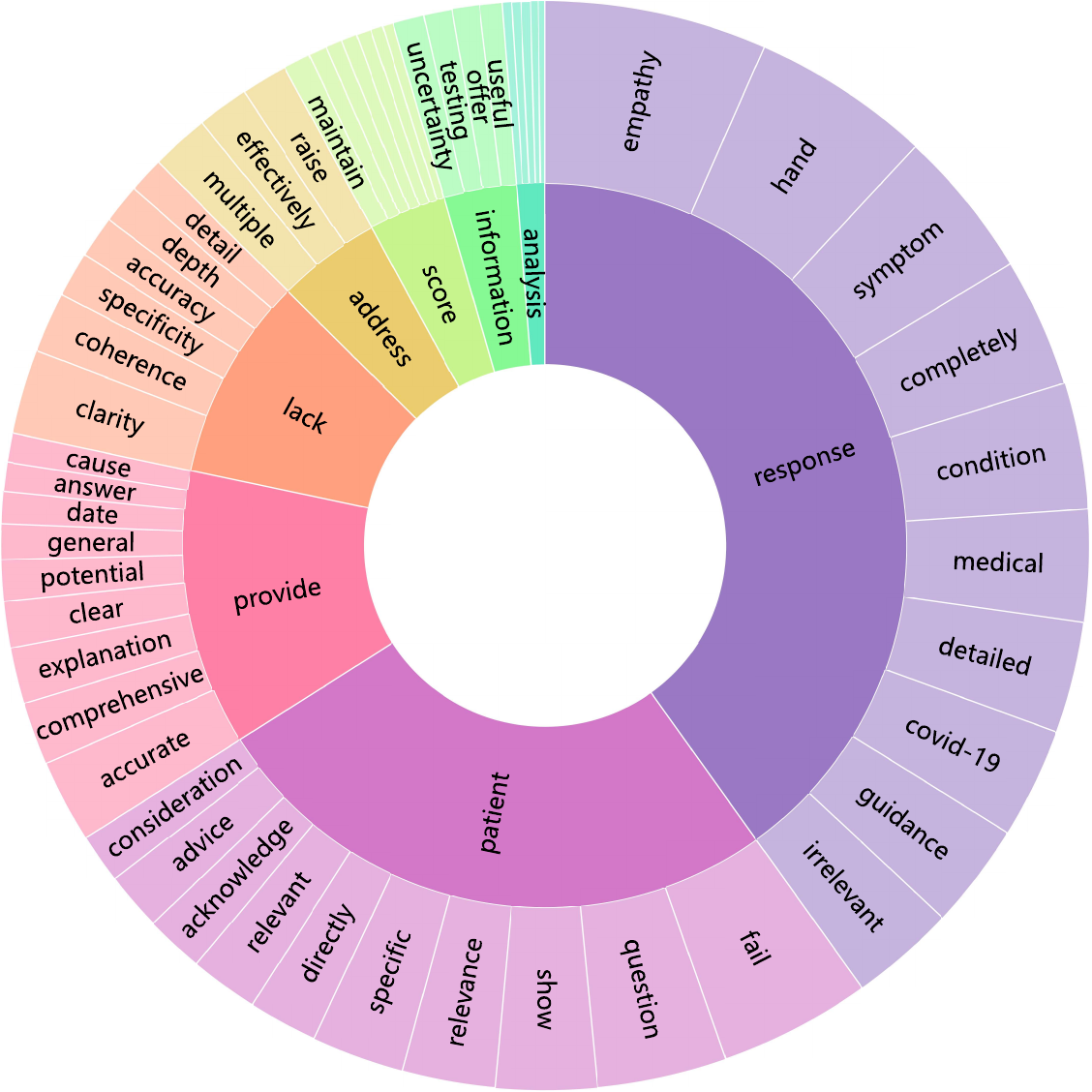}
    \caption{Word co-occurrence graph of collected evaluations in Conclusion, more graphs can be seen in Appendix~\ref{Words Co-occurrence Graph}.}
    \label{fig:conclusion co-occurrence}
    \vspace{-3mm}
\end{figure}

\begin{table}[htbp]
  \centering
  \resizebox{0.9\linewidth}{!}{
    \begin{tabular}{lccr}
    \Xhline{1.0pt}
    Stage & Data &  Learning Rate & Warmup Steps\\
    \hline
    IT & Text-only & \num{2e-5} & 48 \\
    Efficient-RTDPO & Text-only & \num{2e-6} & 15 \\
    IT & Image--Text & \num{2e-5} & 18 \\
    Efficient-RTDPO & Image--Text & \num{1e-6} & 6 \\
    \Xhline{1.0pt}
    \end{tabular}
    }
    \caption{Hyperparameters in different training stages. IT denotes instruction tuning.}
    \label{tab:training stage}
    \vspace{-4mm}
\end{table}

\begin{table}[!ht]
  \centering
  \resizebox{\linewidth}{!}{
    \begin{tabular}{llrrrr}
    \Xhline{1pt}
    \textbf{Dataset} & \textbf{Stage} & \textbf{Conclusion} & \textbf{EXP}   & \textbf{MKC}   & \textbf{PQR} \\
    \hline
    \multirow{3}[0]{*}{
\begin{minipage}{2.6cm}
MedDialogue-\\EN
\end{minipage}} & IT    & 15,210 & 15,694 & 15,664 & 15,760 \\
          & E-RTDPO & 5,376 & 5,088 & 5,178 & 4,890 \\
          & Test  & 1,020 & 1,020 & 1,020 & 1,020 \\
    \hline
    \multirow{3}[0]{*}{MedText} & IT    & 36,805 & 37,206 & 37,217 & 37,271 \\
          & E-RTDPO & 12,606 & 12,402 & 12,369 & 12,207 \\
          & Test  & 1,020 & 1,020 & 1,020 & 1,020 \\
    \hline
    \multirow{3}[0]{*}{MedBench} & IT    & 1,386 & 1,407 & 1,451 & 1,410 \\
          & E-RTDPO & 519   & 468   & 336   & 459 \\
          & Test  & 174   & 174   & 174   & 174 \\
    \hline
    \multirow{3}[0]{*}{\begin{minipage}{2.6cm}
Meadow-\\Wikidoc
\end{minipage}} & IT    & 6,608 & 7,117 & 7,149 & 7,156 \\
          & E-RTDPO & 2,475 & 2,370 & 2,274 & 2,253 \\
          & Test  & 825   & 825   & 825   & 825 \\
    \hline
    \multirow{3}[0]{*}{Path-VQA } & IT    & 9,870 & 11,247 & 11,211 & 11,316 \\
          & E-RTDPO & 3,621 & 3,573 & 3,681 & 3,366 \\
          & Test  & 1,000 & 1,000 & 1,000 & 1,000 \\
    \hline
    \multirow{3}[0]{*}{SLAKE} & IT    & 10,057 & 11,795 & 11,808 & 11,902 \\
          & E-RTDPO & 3,684 & 3,813 & 3,774 & 3,492 \\
          & Test  & 1,000 & 1,000 & 1,000 & 1,000 \\
    \hline
    \multirow{3}[0]{*}{VQA-RAD} & IT    & 3,150 & 3,685 & 3,688 & 3,741 \\
          & E-RTDPO & 1,158 & 1,227 & 1,218 & 1,059 \\
          & Test  & 394   & 394   & 394   & 394 \\
    \Xhline{1pt}
    \end{tabular}
    }
    \caption{Dataset division of different stages, where IT denotes instruction tuning and E-RTDPO denotes Efficient-RTDPO.}
    \label{tab:set split}
    \vspace{-3mm}
\end{table}

\section{Experiments}
\subsection{Experimental Settings}
We train our ACE-$M^3$ model using 2 NVIDIA A100 80GB GPUs. As shown in Table~\ref{tab:training stage}, the training procedure includes four stages. In every stage, our model undergoes one training epoch with a batch size of 128. RMSprop optimization with a Warmup-Decay learning rate schedule and Flash attention 2~\cite{dao2022flashattention,daoflashattention} is employed during the entire training phase. Moreover, we adopt a greedy decoding strategy to avoid randomness. For the vision encoder, we choose ViT-L/14~\footnote{https://huggingface.co/openai/clip-vit-large-patch14}~\cite{radford2021learning} across all multimodal experiments. Details of the training and test set split are given in Table~\ref{tab:set split}.

\subsection{Baselines}
We employ a range of both open-source and closed-source multimodal models as evaluator baselines. Among the open-source models, we select the LLaVA-v1.6 series models~\cite{liu2024llavanext}, as they stand out for their state-of-the-art results across various multimodal benchmarks\footnote{https://llava-vl.github.io/blog/2024-05-10-llava-next-stronger-llms/}. As for closed-source models, we choose the Gemini series models and GPT-4-Turbo (gpt-4-turbo-2024-04-09) as comparative baselines. Since none of these models have undergone instruction fine-tuning, we utilize the one-shot prompting method to standardize their output formats for comparison purposes. Additionally, given the current lack of instruction fine-tuned multimodal evaluation models, we opt to use the fine-tuned PandaLM model for comparison with our model on text-only modality data. 

\begin{table*}[htbp]
  \centering
  \resizebox{0.94\linewidth}{!}{
    \begin{tabular}{lrrrrrrrrrrr}
    \Xhline{1.0pt}
    \multirow{2}[3]{*}{\textbf{Model}} & \multirow{2}[3]{*}{Conclusion} & \multicolumn{4}{c}{EXP} & \multicolumn{3}{c}{MKC} & \multicolumn{3}{c}{PQR} \\
     \cmidrule(lr){3-6} \cmidrule(lr){7-9} \cmidrule(lr){10-12}
     &  & CR & LA & TE & EI & FA & UI & HU & CA & RPC & AMC \\
    \hline
    Random &  \multicolumn{11}{c}{36.00} \\
    \hline
    \hline
     \rowcolor{gray!25}\multicolumn{12}{l}{\textit{Text-only Data}} \\
    LLaVA-v1.6-Mistral-7b & 59.69  & 58.28  & 58.84  & 53.77  & 57.52  & 56.37  & 54.46  & 49.75  & 59.03 & 57.12 & 57.78 \\
    LLaVA-v1.6-Vicuna-7b &  31.46  & 57.03  & 52.32  & 54.49  & 54.49  & 52.88  & 49.46  & 45.21  & 53.93 & 53.01 & 51.69 \\
    LLaVA-v1.6-Vicuna-13b & 50.64  & 62.06  & 59.43  & 52.68  & 61.34  & 58.54  & 56.60  & 43.60  & 60.94 & 57.29 & 57.29 \\
    Gemini-1.0-Pro  & 71.57  & \underline{69.23}  & 63.54  & 60.97  & 68.31  & 57.12  & 57.09  & 49.29  & \underline{69.92}  & 66.37  & 69.04  \\
    Gemini-1.5-Flash & 2.90  & 10.66  & 10.33  & 10.33  & 10.63  & 15.27  & 14.38  & 13.36  & 13.72  & 13.46  & 14.38  \\
    Gemini-1.5-Pro & 27.64  & 2.83  & 2.80  & 2.73  & 2.93  & 4.38  & 4.64  & 4.38  & 24.22  & 22.77  & 24.84  \\
    GPT-4-Turbo & \textbf{76.79}  & 68.86  & \textbf{66.08}  & \textbf{74.94}  & \textbf{72.15}  & \textbf{71.97}  & \textbf{69.70}  & \textbf{70.96}  & \textbf{73.99}  & \textbf{73.74}  & \textbf{76.26}  \\
    PandaLM  & 70.88   &  \multicolumn{10}{c}{/}  \\
    Ours (ACE-$M^3$) & \underline{72.59}  & \textbf{69.46}  & \underline{63.67}  & \underline{70.52}  & \underline{68.38}  & \underline{59.26}  & \underline{59.39}  & \underline{61.96}  & 69.46  & \underline{67.98}  & \underline{72.49}  \\
    \hline
    \hline
    \rowcolor{gray!25}\multicolumn{12}{l}{\textit{Image--Text Data}} \\
    LLaVA-v1.6-Mistral-7b & 65.96  & 33.12  & 29.82  & 34.17  & 23.31  & 54.34  & 48.96  & 49.87  & 58.69 & 46.32 & 54.64 \\
    LLaVA-v1.6-Vicuna-7b & 0.50  & 13.45  & 13.62  & 12.53  & 9.77  & 15.04  & 15.04  & 15.79  & 16.96 & 13.49 & 17.17 \\
    LLaVA-v1.6-Vicuna-13b & 9.15  & 24.44  & 23.68  & 22.26  & 23.31  & 31.50  & 26.90  & 30.87  & 40.27 & 30.53 & 31.50 \\
    Gemini-1.0-Pro-Vision & 75.86  & 71.47  & 69.47  & 70.09  & 64.54  & 53.80  & 46.91  & 47.45  & 71.51  & 53.43  & 56.10  \\
    Gemini-1.5-Flash & 64.37  & 60.78  & 56.77  & 66.08  & 49.46  & 54.34  & \underline{53.68}  & \underline{55.18}  & 70.26  & 48.71  & 54.68  \\
    Gemini-1.5-Pro &  30.74  & 5.68  & 5.26  & 5.81  & 5.30  & 15.58  & 16.33  & 15.20  & 30.49  & 22.60  & 25.65  \\
    GPT-4-Turbo & \underline{78.57}  & \underline{79.73}  & \underline{72.30}  & \textbf{81.42}  & \underline{75.34}  & \underline{58.45}  & 50.34  & 50.34  & \underline{79.39}  & \underline{64.86} & \underline{59.46} \\
    Ours (ACE-$M^3$) & \textbf{82.71}  & \textbf{80.41}  & \textbf{73.73}  & \underline{79.11}  & \textbf{78.03}  & \textbf{67.13}  & \textbf{63.95}  & \textbf{62.70}  & \textbf{83.58}  & \textbf{68.25}  & \textbf{67.13}  \\
    \Xhline{1.0pt}
    \end{tabular}%
  }
  \caption{Main experimental results. The bolded numbers indicate the best performance, while the underlined numbers represent the second-best performance. We run our model three times and report the average Accuracy and the variance of all metrics is smaller than 0.01. More detailed results can be found in Appendix~\ref{Detailed Main Results} and~\ref{score distribution}.}
  \label{tab:main results}
  \vspace{-1mm}
\end{table*}

\subsection{Metrics}
We use Accuracy as the metric for automatic evaluation, which measures the consistency between the relative magnitude of the scoring outcomes generated by ACE-$M^3$  for medical models and the relative magnitude of labeled scores. Human evaluation is  conducted as well, which is discussed later in Section~\ref{Reliablity of model}.

\subsection{Main Results}
As shown in Table~\ref{tab:main results}, our model outperforms all other models on the image--text data, especially surpassing GPT-4-Turbo by 5.3\% relative improvement in terms of Accuracy from the conclusion evaluation, indicating the effectiveness of our ACE-$M^3$ model. Surprisingly, even with one-shot prompting, the LLaVA models built upon the Vicuna series perform extremely unsatisfactorily on the image--text data due to incorrect output formatting, and the same issue also occurs with the Gemini-1.5-Flash model on the text-only data. This indicates a lack of instruction-following capability and highlights the necessity of developing specialized evaluation models for specific domains.

\begin{table}[htbp]
  \centering
  \resizebox{0.98\linewidth}{!}{
    \begin{tabular}{lccc}
    \Xhline{1.0pt}
 \textbf{Model}   & Text-only Data & Image-Text Data & Average \\
    \hline
    Ours (ACE-$M^3$)  & 72.59 & 82.71 & 77.05 \\
   -\emph{w/o} Reward Token & 72.46  & 80.99  & 76.22  \\
    -\emph{w/o} DPO  & 71.17 & 82.58 & 76.20 \\
    \Xhline{1.0pt}
    \end{tabular}
  }
    \caption{Ablation results of different training strategies on two kinds of data.}
    \label{tab:ablation study}
    \vspace{-3mm}
\end{table}

\subsection{Ablation Studies}
\subsubsection{Impact of RTDPO}
As shown in Table~\ref{tab:ablation study}, we conduct ablation studies on both kinds of data to quantify the contributions of different strategies in the training of ACE-$M^{3}$. The two components contribute to different kinds of data: The removal of our direct preference optimization variant causes more degradation on the text-only data, while the ablation of the reward token leads to a decrease on the multimodal data. More detailed ablation results are shown in Appendix~\ref{Detailed Ablation Results}.

\subsubsection{Influence of Frozen Layers}
To investigate the impact of varying frozen layers of the LLM, we conducted experiments by training ACE-$M^3$ with different frozen layers.

\begin{figure}[htbp]
    \centering
\includegraphics[width=0.96\linewidth]{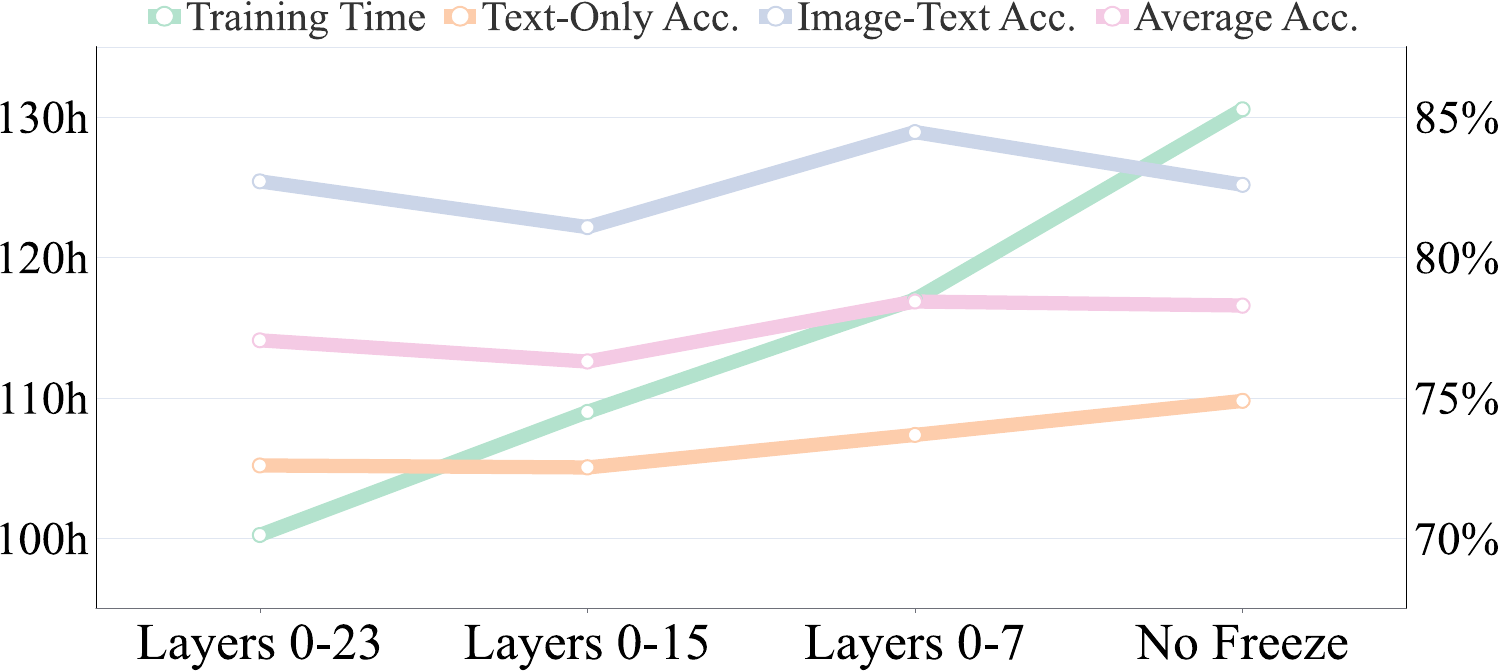}
    \caption{Influence of frozen layers on time and evaluation accuracy.}
    \label{fig:frozen layers}
\end{figure}

As depicted in Figure~\ref{fig:frozen layers}, the training time continuously increases with the number of frozen layers decreasing. Compared to freezing the parameters of the first 24 layers, not freezing any parameters results in an approximate 30\% increase in training time. However, with the number of frozen layers decreasing, the model's evaluation accuracy does not improve significantly, which proves the effectiveness of freezing lower layers to the trade-off between training time and evaluation accuracy. The influence of frozen layers on each dataset can be found in Appendix~\ref{Detailed Ablation Results}.

\begin{tcolorbox}[
    fonttitle = \fontsize{10pt}{12pt}\selectfont,
    colback = white,
    colbacktitle = black!85!black, enhanced,
    attach boxed title to top left={yshift=-4mm, xshift=3mm},
    title=Conclusion Evaluation Case,
    left=0mm,
    right=0mm,
    bottom=0mm,
    fontupper=\fontsize{7.1pt}{8.1pt}\selectfont,
    fontlower=\fontsize{7.1pt}{8.1pt}\selectfont,
    ]
    \vspace{2mm}
    \begin{tabular}{ll}
    \multirow{1}[0]{*}{{\makecell[{{>{\raggedright}}p{1.2cm}}]{Image:}}}  & \resizebox{0.2\linewidth}{!}{
            \includegraphics{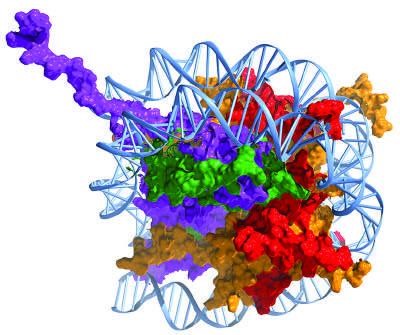}
        } \\
       {{\makecell[{{>{\raggedright}}p{1.1cm}}]{User:}}}  & \multirow{2}{*}{\makecell[{{>{\raggedright}}p{5.6cm}}]{What are positively charged, thus allowing the compaction of the negatively charged dna?}} \\
       &\\
       &\\
       {\makecell[{{>{\raggedright}}p{1.1cm}}]{Model 1:}}  & \multirow{3}{*}{\makecell[{{>{\raggedright}}p{5.6cm}}]{The positively charged ions are the cations, which are attracted to the negatively charged DNA, allowing the compaction of the negatively charged dna.}} \\
       &\\
       &\\
       &\\
       {\makecell[{{>{\raggedright}}p{1.1cm}}]{Model 2:}}  & \multirow{4}{*}{\makecell[{{>{\raggedright}}p{5.6cm}}]{The image shows that histone proteins have positively charged amino acid residues on their surface, which helps them to interact with and compact the negatively charged DNA.}}\\
       &\\
       &\\
    \end{tabular}
    \tcblower
    \definecolor{lightblue}{RGB}{50,191,227}
    \definecolor{lightred}{RGB}{233,91,91}
    \textbf{Reference Evaluation}:\\
    Analysis:\\Response 1: Response 1 provides a clear explanation of how cations interact with negatively charged DNA for compaction. However, it lacks depth in addressing the specific role of histone proteins.\\Response 2: Response 2 offers a detailed explanation of how histone proteins with positively charged amino acid residues interact with and compact negatively charged DNA, directly addressing the patient's question.\\\\Final Score:\\Response 1: 3\\Response 2: 5\\
        \\
    \textbf{Ours (ACE-$M^3$)}:\\
    Analysis:\\Response 1: Response 1 gives a straightforward explanation of the interaction between cations and negatively charged DNA for compaction. \textcolor{lightblue}{Nevertheless, it doesn't delve deeply into the specific function of histone proteins.} \\Response 2: Response 2 provides a comprehensive explanation of how histone proteins, which have positively charged amino acid residues, interact with and compact negatively charged DNA, directly answering the patient's question.\\\\Final Score:\\Response 1: 3\\Response 2: 5 \\
       \\
    \textbf{LLaVA-v1.6-Vicuna-13b}:\\
    Analysis:\\Response 1: Response 1 accurately suggests that cations are the primary agents in DNA compaction \textcolor{lightred}{without mentioning the critical role of histone proteins.} This response is overly simplistic and misses the key aspect of histone involvement.\\Response 2: Response 2 is less clear and \textcolor{lightred}{somewhat confusing.} It doesn't mention histone proteins and their interaction with DNA and lacks precision and depth, making it a less satisfactory answer.\\\\Final Score:\\\textcolor{lightred}{Response 1: 5}\\\textcolor{lightred}{Response 2: 4} \\
       \\
    \textbf{Gemini-1.0-Pro-Vision}:\\
    Analysis:\\Response 1: Response 1 explains how positively charged ions (cations) interact with negatively charged DNA to facilitate compaction. \textcolor{lightblue}{However, it does not specifically mention histone proteins, which play a crucial role in DNA compaction.}\\Response 2: Response 2 provides a more comprehensive answer by explaining that histone proteins, with their positively charged amino acid residues, interact with and compact negatively charged DNA. This response is more detailed and directly addresses the mechanism of DNA compaction.\\\\Final Score:\\\textcolor{lightred}{Response 1: 4}\\Response 2: 5
\end{tcolorbox}

\section{Discussions}

\subsection{Case Study}
The conclusion evaluation case shown in the above box demonstrates how our model first analyzes the two responses and highlights that the response of Model 1 lacks explanations of key information, while Model 2's response provides detailed and comprehensive information. Thus, our model assigns the scores of 3 and 5 to the two models, respectively, which appear reasonable and accurate. Meanwhile, Gemini-1.0-Pro-Vision's evaluation of the two responses is similar to our model, but the scores assigned to both responses are less accurate. The LLaVA-v1.6-Vicuna-13b model's analysis of Model 1’s response appears to be somewhat justified. Nonetheless, its analysis of Model 2's response is entirely erroneous, and the scores that it assigns are incorrect. More detailed examples can be seen in Appendix~\ref{more examples}.

\subsection{Is the Instruction Dataset Reliable?}
The reliability of the instruction dataset is paramount for ensuring the validity and effectiveness of any model trained on it. We discuss it from the following two points.

\paragraph{Detailed and Reliable Criteria} 
The criteria (proposed in Section~\ref{Reliable Evaluation Criteria}) used to guide the data collection process are meticulously designed and detailed. These criteria are established to maintain consistency and accuracy in the data, ensuring that each entry meets the predefined standards. By adhering to these criteria, we minimize the risk of including erroneous or irrelevant data in the dataset.

\paragraph{Reference-guided Evaluation} 
While constructing large-scale datasets through manual annotation is both time-consuming and costly, ChatGPT is proven to be a competitive evaluator compared to human judgments, especially with reference answers~\cite{wang2023chatgpt}. Therefore, the current standard practice leverages the capabilities of large language models to build training datasets~\cite{li2024leveraging} and previous text-only evaluation models such as PandaLM~\cite{wang2024pandalm} and Auto-J~\cite{ligenerative} are built upon synthetic LLM data.

\paragraph{Two-step Verification} We employ two-step verification including automatic format checking and human sampling content verification. The statistics in Table~\ref{tab:statistics} indicate that our dataset is reliable.

\begin{table*}[htbp]
  \centering
  \resizebox{0.98\linewidth}{!}{
    \begin{tabular}{lrrrrrrrrrrr}
    \Xhline{1.0pt}
    \multicolumn{1}{l}{\multirow{2}[1]{*}{\textbf{Encoder}}} & \multicolumn{1}{c}{\multirow{2}[1]{*}{Conclusion}} & \multicolumn{4}{c}{EXP} & \multicolumn{3}{c}{MKC} & \multicolumn{3}{c}{PQR} \\
    \cmidrule(lr){3-6} \cmidrule(lr){7-9} \cmidrule(lr){10-12}
          & & \multicolumn{1}{c}{CR} & \multicolumn{1}{c}{LA} & \multicolumn{1}{c}{TE} & \multicolumn{1}{c}{EI} & \multicolumn{1}{c}{FA} & \multicolumn{1}{c}{UI} & \multicolumn{1}{c}{HU} & \multicolumn{1}{c}{CA} & \multicolumn{1}{c}{RPC} & \multicolumn{1}{c}{AMC} \\
    \hline
    CLIP  & 82.71 & 80.41 & 73.73 & 79.11 & 78.03 & 67.13 & 63.95 & 62.7  & 83.58 & 68.25 & 67.13 \\
    PubMedCLIP & 83.00  & 80.12  & 73.89  & 79.41  & 77.74  & 67.42  & 64.04  & 62.91  & 83.96  & 67.92  & 66.88  \\
    BiomedCLIP & 83.29  & 80.58  & 73.77  & 79.37  & 78.53  & 67.46  & 64.16  & 63.07  & 83.58  & 68.55  & 67.63  \\
    \Xhline{1.0pt}
    \end{tabular}
    }
    \caption{Accuracy rate of each aspect using different encoding techniques.}
    \label{tab:different encoder}
\end{table*}

\subsection{Is ACE-$M^3$ Reliable?}
\label{Reliablity of model}
ACE-$M^3$ has been proven effective with the automatic evaluations in the main results in Table~\ref{tab:main results}. Additionally, we invited the aforementioned annotators to annotate 200 samples (120 samples from the text-only test set and 80 samples from the image--text test set with annotation details in Appendix~\ref{Annotation Details}). As shown in Figure~\ref{fig:correlation}, the evaluation results of our ACE-$M^3$ model for various medical models largely overlap with the assessments of human annotators, indicating a high reliability of our model. Furthermore, traditional metrics such as BLEU and METEOR diverge substantially from human preferences.

\begin{figure}[htbp]
    \centering
\includegraphics[width=0.92\linewidth]{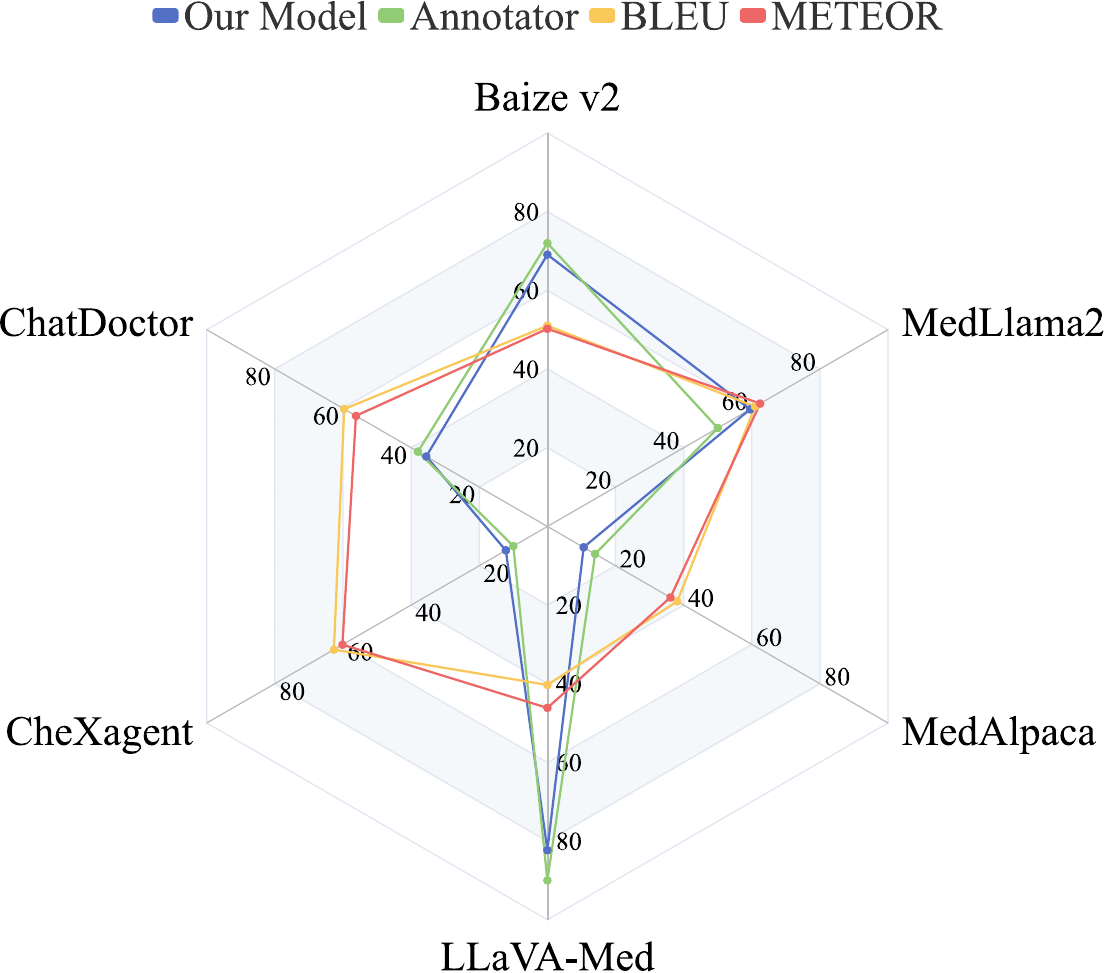}
    \caption{Win rate judged by our ACE-$M^3$ model, human annotator, and traditional metrics.}
    \label{fig:correlation}
\end{figure}


\subsection{Influences of Vision Encoders}
We conduct experiments over different encoding techniques such as PubMedCLIP~\cite{EslamiDeMelo2023PubMedCLIP} and 
BiomedCLIP~\cite{zhang2023biomedclip} to inspect the influence of the selection of encoders. As depicted in Table~\ref{tab:different encoder}, domain-specific encoders can offer certain improvements, indicating their enhanced ability to extract relevant medical features from images effectively.


\subsection{Bias}
In this section, we conduct experiments and comparisons to investigate potential biases, including position bias, verbosity bias, and symmetry bias that may exist in the ACE-$M^3$ model when used as an evaluator.

\begin{table}[htbp]
  \centering
  \resizebox{\linewidth}{!}{
    \begin{tabular}{lccc}
    \Xhline{1.0pt}
    \textbf{Model} & Accuracy$_{first}$ & Accuracy$_{second}$ & Difference ($\downarrow$)\\
    \hline
    Ours (ACE-$M^3$)  & 83.62  & 84.05  & \textbf{0.42}  \\
    PandaLM  & 83.52  & 77.41  & 6.11  \\
    GPT-4-Turbo & 81.88  & 85.28  & 3.39  \\
    Gemini-1.0 & 82.19  & 78.13  & 4.06 \\
    Gemini-1.5-Flash & 34.02  & 31.59  & 2.43 \\
    Gemini-1.5-Pro & 35.37  & 28.99  & 6.38 \\
    LLaVA-v1.6-Mistral-7b & 65.46  & 68.05  & 2.59 \\
    LLaVA-v1.6-Vicuna-7b & 26.96  & 12.12  & 14.84 \\
    LLaVA-v1.6-Vicuna-13b & 33.19  & 33.71  & 0.52 \\
    \Xhline{1.0pt}
    \end{tabular}
  }
    \caption{Evaluation accuracy when better response is in first or second position.}
    \label{tab:position bias}
\end{table}

\paragraph{Position Bias}
Position bias occurs when an MLLM serving as an evaluator prefers answers in certain positions over others.
We measure the model's preference for different positions by analyzing the differences in accuracy at various positions. As shown in Table~\ref{tab:position bias}, the accuracy difference between the two positions in our ACE-$M^3$ model is significantly smaller than that of PandaLM and GPT-4-Turbo, which indicates that our model exhibits less positional bias.

\paragraph{Verbosity Bias}
Verbosity bias refers to whether the evaluation model prefers longer responses or shorter ones. As shown in Table~\ref{tab:verbosity bias}, our model prefers longer responses than PandaLM and GPT-4-Turbo. The reason is that in real-life scenarios, doctors give concise responses tailored to the patient's situation. However, when patients ask brief questions to models, the latter have to generate longer content by listing various solutions for different scenarios.

\begin{table}[htbp]
  \centering
  \resizebox{\linewidth}{!}{
    \begin{tabular}{lccc}
    \Xhline{1.0pt}
    \textbf{Model} & Accuracy$_{longer}$ & Accuracy$_{shorter}$ & Difference ($\downarrow$)\\
    \hline
    Ours (ACE-$M^3$)  & 91.23  & 61.97  & 29.26  \\
    PandaLM  & 89.20  & 66.18  & 23.02  \\
    GPT-4-Turbo	& 87.84 & 68.66 & 19.18 \\
    Gemini-1.0 & 84.22  & 67.64  & 16.58 \\
    Gemini-1.5-Flash & 41.19  & 7.42  & 33.77 \\
    Gemini-1.5-Pro & 33.29  & 28.19  & \textbf{5.10} \\
    LLaVA-v1.6-Mistral-7b & 68.97  & 60.38  & 8.58 \\
    LLaVA-v1.6-Vicuna-7b & 15.00  & 31.86  & 16.86 \\
    LLaVA-v1.6-Vicuna-13b & 29.22  & 46.12  & 16.90 \\
    \Xhline{1.0pt}
    \end{tabular}
  }
    \caption{Evaluation accuracy when the length of the better response is longer or shorter.}
    \label{tab:verbosity bias}
\end{table}


\paragraph{Symmetry Bias}
Symmetry bias denotes whether the evaluation results of a model change if the positions of two responses are swapped. During experiments, we find that 9.52\% of the samples evaluated by our model exhibit symmetry bias compared to 16.37\% in PandaLM and 12.88\% in GPT-4-Turbo, which evinces better robustness of our model.


\section{Related Work}
Evaluation methodologies for multimodal large language models (MLLMs) typically fall into two fundamental categories: closed-set and open-set evaluations~\cite{yin2024survey}.

For closed-set evaluations, some benchmarks have been proposed to evaluate the capability of medical MLLMs, such as Path-VQA~\cite{he2020pathvqa}, SLAKE~\cite{9434010}, and VQA-RAD~\cite{lau2018dataset}. In contrast, when the questions in the benchmark are open-ended, traditional automated metrics, such as F1-score~\cite{chinchor1993muc}, BLEU~\cite{papineni2002bleu}, and ROUGE~\cite{2004ROUGE}, are utilized for evaluation. However, most traditional automated metrics assess the effectiveness of models solely at the lexical level, which is inadequate for more complex generation tasks, due to their failure to consider semantics and poor alignment with human judgments. Therefore, it remains a challenging task to evaluate the open-ended QA performance of MLLMs with benchmarks~\cite{zheng2024judging}.

In principle, it is possible to conduct human evaluations on the entire dataset for open-set evaluation~\cite{xu2023medgpteval}. However, it is highly impractical to solicit humans to evaluate the effectiveness of models at a larger scale, as it requires a substantial allocation of resources, including both time and money. Meanwhile, MLLMs continue to advance at a rapid pace, but the progress on automated evaluation methods to assess their performance has lagged. Although GPT-4 and Gemini can aid automated assessments~\cite{nori2023capabilities, li2024leveraging, wang2023chatgpt}, they remain suboptimal options due to their proprietary nature and lack of reproducibility. Open-sourced evaluation models such as PandaLM have been proposed for generic text-only tasks, but they are unable to perform multimodal evaluations.

\section{Conclusion}
In this paper, we propose an automated multimodal evaluation model ACE-$M^3$, along with an instruction dataset utilized to train the model, which can facilitate the automatic evaluation of MLLMs in the medical field. Specifically, we use medical visual question-answer data and detailed evaluation criteria to collect evaluation results from the ChatGPT model, and train the branch-merged architecture evaluation model ACE-$M^3$ by utilizing the collected datasets.

We further propose an Efficient-RTDPO training strategy that comprises two main components. One component utilizes the advanced RTDPO training method to precisely enhance the model's inherent ability to generate more accurate and detailed evaluation content as well as reliable scores. The other component involves freezing the parameters in the lower layers of LLMs to improve training efficiency without significantly compromising the accuracy of the evaluations. The model's
performance and training cost benefit from the two training techniques in comparative and ablation experiments.

\section*{License}
The dataset and models used in this paper are open-sourced or permitted to be used in the science research area. The ACE-$M^3$ model in this paper is trained by using the open-sourced data and models, which leads to the restriction that ACE-$M^3$ should only be used for research purposes.

\section*{Ethics Statement}
The training data utilized and constructed in this article is both publicly available and anonymized, thus ensuring that no personal privacy issues are involved. We caution that the ACE-$M^3$ model is primarily designed to help gauge the performance of medical MLLMs. It is not intended to prove the medical MLLMs' suitability or effectiveness for genuine real-world deployment.

\section*{Limitations}
Despite our evaluation model ACE-$M^3$ demonstrating high accuracy in automated assessments and showing a strong correlation with human evaluation results, it still has some shortcomings. We randomly sample 20 erroneous instances from the model's evaluation outcomes for further analysis and identify the following issues: (1) 2 examples exist of misattribution of two responses' content, (2) incorrect medical knowledge leads to incorrect results in 10 cases, and (3) unsupported ratings or hallucination in 8 cases. Therefore, the model cannot ensure that the evaluation results of the model are fully aligned with human preferences.

\section*{Acknowledgements}
This work was supported by NSFC grant (No. 62136002 and 62477014), Ministry of Education Research Joint Fund Project (8091B042239), and Shanghai Trusted Industry Internet Software Collaborative Innovation Center.

\bibliography{custom}
\appendix
\clearpage
\onecolumn

\section{Detailed Criteria for Three Sub-domains}
\label{Detailed Criteria}
\begin{tcolorbox}[breakable, colback=gray!20, width=\textwidth, fontupper=\fontsize{8pt}{9pt}\selectfont]
\textbf{Expression}:\\
Clarity of Response: Assess how well the model expresses information. Evaluate the responses for coherence, logical flow, and clarity. \\
0: Response is entirely unclear and confusing. \\
1: Major clarity issues, difficult to understand. \\
2: Some clarity issues, but main points are discernible. \\
3: Clear and logically structured response. \\
4: Very clear, concise, and well-structured. \\
5: Exceptionally clear, concise, and well-structured. \\

Language Appropriateness: Evaluate whether the model uses language suitable for the target audience (patients). Check for jargon, complex terms, or overly simplistic language. \\
0: Inappropriate language or excessive use of jargon. \\
1: Major issues with language appropriateness. \\
2: Some inappropriate language, but main message is understandable. \\
3: Language is generally suitable for the target audience. \\
4: Language is highly suitable and engaging. \\
5: Language is both entirely appropriate and engaging. \\

Tone and Empathy: Assess the model's ability to convey information in a compassionate and empathetic manner. Ensure that responses are sensitive to the patient's emotional state. \\
0: Lack of empathy, insensitive response. \\
1: Major issues with empathy, highly insensitive. \\
2: Some attempt at empathy, but it could be improved. \\
3: Empathetic and sensitive response. \\
4: Highly empathetic and sensitive. \\
5: Exceptional empathy, demonstrating a deep understanding of the patient's emotions. \\

Expression Integrity: Evaluate the overall integrity of the model's expression, taking into account how well it maintains consistency and coherence throughout the response. \\
0: Response lacks any semblance of coherence and consistency. \\
1: Major issues with expression integrity, making the response disjointed. \\
2: Some lapses in expression integrity, but the overall message is still discernible. \\
3: Expression is generally consistent and coherent. \\
4: Highly consistent expression with minimal lapses in coherence. \\
5: Exceptionally consistent expression, demonstrating a seamless and coherent flow of information. \\

\textbf{Medical Knowledge Correctness:}\\
Factual Accuracy: Evaluate the responses for the accuracy of medical information. Cross-reference responses with authoritative medical sources to ensure correctness. \\
0: Information provided is entirely incorrect. \\
1: Major inaccuracies present. \\
2: Several inaccuracies present. \\
3: Generally accurate, with minor errors. \\
4: Mostly accurate, with very minor exceptions. \\
5: Information is entirely accurate. \\

Up-to-date Information: Check if the model provides information that is current and reflects the latest medical knowledge. \\
0: Information is outdated or obsolete. \\
1: Major outdated information. \\
2: Some outdated information. \\
3: Mostly up-to-date, with minor exceptions. \\
4: Information is current and mostly reflects the latest medical knowledge. \\
5: Information is entirely current and reflects the latest medical knowledge. \\

Handling Uncertainty: Assess how the model deals with ambiguous or uncertain situations. It should communicate uncertainty when appropriate and avoid giving misleading information.  \\
0: Model consistently provides misleading information. \\
1: Major difficulty in handling uncertainty. \\
2: Some difficulty in handling uncertainty. \\
3: Adequate acknowledgment of uncertainty. \\
4: Highly adept at handling uncertainty with transparent communication. \\
5: Exceptional handling of uncertainty, with transparent communication.\\
\\
\textbf{Patient Question Relevance:}\\
Context Awareness: Evaluate the model's understanding of the context of the patient's question. Ensure that responses address the specific concerns raised by the patient. \\
0: Responses consistently lack relevance to the context. \\
1: Major relevance issues, with little connection to the context. \\
2: Some relevance issues, but key points addressed. \\
3: Relevant responses tailored to the context. \\
4: Highly relevant responses, demonstrating good understanding of the context. \\
5: Responses demonstrate exceptional understanding of the context. \\

Relevance to Patient's Condition: Assess whether the model tailors responses to the individual patient's health condition, if available. \\
0: Responses show no consideration for the patient's condition. \\
1: Major issues with considering the patient's condition. \\
2: Limited consideration, with some relevance. \\
3: Consideration of the patient's condition evident in responses. \\
4: Responses show a high degree of consideration for the patient's condition. \\
5: Responses are highly tailored to the individual patient's health condition. \\

Addressing Multiple Concerns: Evaluate the model's ability to handle questions that involve multiple medical concerns, providing comprehensive and relevant information.  \\
0: Model struggles to address multiple concerns coherently. \\
1: Major difficulty in addressing multiple concerns. \\
2: Some attempts to address multiple concerns, with limitations. \\
3: Competent handling of questions with multiple concerns. \\
4: Very competent at addressing and integrating multiple concerns in responses. \\
5: Exceptional ability to address and integrate multiple concerns in responses.
\end{tcolorbox}

\section{Prompts}
\subsection{Prompt for Response Collection}
\label{Prompt for Response Collection}
\begin{tcolorbox}[colback=gray!20,  width=\textwidth, fontupper=\fontsize{8pt}{9pt}\selectfont]
\textbf{Text-only models:}\\
Please answer the following question faithfully.\\
\\
\#\#\# Question: \{question\}\\
\\
\#\#\# Answer: \\
\\
\textbf{Image-Text models:}\\
Please answer the following question based on the image provided.\\
\\
<image>\\
\{question\}\\
\\
Response: 
\end{tcolorbox}

\subsection{Prompt for Evaluation Collection}
\label{Prompt for Evaluation Collection}
\begin{tcolorbox}[colback=gray!20, width=\textwidth, breakable, fontupper=\fontsize{8pt}{9pt}\selectfont]
\textbf{Expression:}\\
\#\#\# Instruction:\\
\#\# Evaluation Criterion: (higher score means better performance)\\
\{expression evaluation criteria\}\\
\\
\#\# Your output should strictly follow the format below and the word between signal \$ represents the content you need to generate:\\
Response 1:\\
Criterion Clarity of Response:\\
Analysis: \$analysis\$\\
Score: \$score\$\\
Criterion Language Appropriateness:\\
Analysis: \$analysis\$\\
Score: \$score\$\\
Criterion Tone and Empathy:\\
Analysis: \$analysis\$\\
Score: \$score\$\\
Criterion Expression Integrity:\\
Analysis: \$analysis\$\\
Score: \$score\$\\
Response 2:\\
Criterion Clarity of Response:\\
Analysis: \$analysis\$\\
Score: \$score\$\\
Criterion Language Appropriateness:\\
Analysis: \$analysis\$\\
Score: \$score\$\\
Criterion Tone and Empathy:\\
Analysis: \$analysis\$\\
Score: \$score\$\\
Criterion Expression Integrity:\\
Analysis: \$analysis\$\\
Score: \$score\$\\
\\
\#\# Below are two doctors' responses to a patient's question about an image. Since you can't see the image, we provide the correct answer to the question for you to refer to. Now you need to pretend that you can see the image 
and analyze each response one by one then give a score between 0-5 to each response about each criterion following the format requirement above.\\
\\
\#\#\# Question:\\
\{question\}\\
\\
\#\#\# Response 1:\\
\{response\_1\}\\
\\
\#\#\# Response 2:\\
\{response\_2\}\\
\\
\#\#\# Reference Answer:\\
\{reference\_answer\}\\
\\
\#\#\# Expression Evaluation:\\

\textbf{Medical Knowledge Correctness:}\\
\#\#\# Instruction:\\
\#\# Evaluation Criterion: (higher score means better performance)\\
\{medical knowledge correctness evaluation criteria\}\\
\\
\#\# Your output should strictly follow the format below and the word between signal \$ represents the content you need to generate:\\
Response 1:\\
Criterion Factual Accuracy:\\
Analysis: \$analysis\$\\
Score: \$score\$\\
Criterion Up-to-date Information:\\
Analysis: \$analysis\$\\
Score: \$score\$\\
Criterion Handling Uncertainty:\\
Analysis: \$analysis\$\\
Score: \$score\$\\
Response 2:\\
Criterion Factual Accuracy:\\
Analysis: \$analysis\$\\
Score: \$score\$\\
Criterion Up-to-date Information:\\
Analysis: \$analysis\$\\
Score: \$score\$\\
Criterion Handling Uncertainty:\\
Analysis: \$analysis\$\\
Score: \$score\$\\
\\
\#\# Below are two doctors' responses to a patient's question about an image. Since you can't see the image, we provide the correct answer to the question for you to refer to. Now you need to pretend that you can see the image and analyze each response one by one then give a score between 0-5 to each response about each criterion following the format requirement above.\\
\\
\#\#\# Question:\\
\{question\}\\
\\
\#\#\# Response 1:\\
\{response\_1\}\\
\\
\#\#\# Response 2:\\
\{response\_2\}\\
\\
\#\#\# Reference Answer:\\
\{reference\_answer\}\\
\\
\#\#\# Medical Knowledge Correctness Evaluation:\\
\\
\textbf{Patient Question Relevance:}\\
\#\#\# Instruction:\\
\#\# Evaluation Criterion: (higher score means better performance)\\
\{patient question relevance evaluation criteria\}\\
\\
\#\# Your output should strictly follow the format below and the word between signal \$ represents the content you need to generate:\\
Response 1:\\
Criterion Context Awareness:\\
Analysis: \$analysis\$\\
Score: \$score\$\\
Criterion Relevance to Patient's Condition:\\
Analysis: \$analysis\$\\
Score: \$score\$\\
Criterion Addressing Multiple Concerns:\\
Analysis: \$analysis\$\\
Score: \$score\$\\
Response 2:\\
Criterion Context Awareness:\\
Analysis: \$analysis\$\\
Score: \$score\$\\
Criterion Relevance to Patient's Condition:\\
Analysis: \$analysis\$\\
Score: \$score\$\\
Criterion Addressing Multiple Concerns:\\
Analysis: \$analysis\$\\
Score: \$score\$\\
\\
\#\# Below are two doctors' responses to a patient's question about an image. Since you can't see the image, we provide the correct answer to the question for you to refer to. Now you need to pretend that you can see the image and analyze each response one by one then give a score between 0-5 to each response about each criterion following the format requirement above.\\
\\
\#\#\# Question:\\
\{question\}\\
\\
\#\#\# Response 1:\\
\{response\_1\}\\
\\
\#\#\# Response 2:\\
\{response\_2\}\\
\\
\#\#\# Reference Answer:\\
\{reference\_answer\}\\
\\
\#\#\# Patient Question Relevance Evaluation:\\

\textbf{Conclusion:}\\
\#\#\# Instruction:\\
\#\# Your output should strictly follow the format below and the word between signal \$ represents the content you need to generate:\\
Analysis:\\
\$analysis\$\\

Final Score:\\
Response 1: \$final\_score\$\\
Response 2: \$final\_score\$\\
\\
\#\# Below are two doctors' responses to a patient's question about an image followed by some sub-domain evaluations of the responses. Since you can't see the image, we provide the correct answer to the question for you to refer to. Now you need to pretend that you can see the image and analyze two responses comprehensively, then give a final score between 0-5 to each response based on the format requirement above.\\
\\
\#\#\# Question:\\
\{question\}\\
\\
\#\#\# Response 1:\\
\{response\_1\}\\
\\
\#\#\# Response 2:\\
\{response\_2\}\\
\\
\#\#\# Reference Answer:\\
\{reference\_answer\}\\
\\
\#\#\# Sub-domain Evaluations:\\
\#\# Sub-domain Patient Question Relevance:\\
\{patient question relevance evaluation result\}\\
\#\# Sub-domain Medical Knowledge Correctness:\\
\{medical knowledge correctness evaluation result\}\\
\#\# Sub-domain Expression:\\
\{expression evaluation result\}\\
\\
\#\#\# Evaluation:
\end{tcolorbox}
\subsection{Prompt for Training}
\label{Prompt for Instruction Tuning}
\begin{tcolorbox}[colback=gray!20, width=\textwidth, breakable, fontupper=\fontsize{8pt}{9pt}\selectfont]
\textbf{Expression:}\\
\#\#\# Instruction:\\
\#\# Evaluation Criterion: (higher score means better performance)\\
\{expression evaluation criteria\}\\
\\
\#\# Below are two doctors' responses to a patient's question about an image. Now you need to analyze each response one by one then give a score between 0-5 to each response about each criterion above.\\
\\
\#\#\# Image And Question:\\
<image>\\
\{question\}\\
\\
\#\#\# Response 1:\\
\{response\_1\}\\
\\
\#\#\# Response 2:\\
\{response\_2\}\\
\\
\#\#\# Expression Evaluation:\\
\{evaluation\}\\

\textbf{Medical Knowledge Correctness:}\\
\#\#\# Instruction:\\
\#\# Evaluation Criterion: (higher score means better performance)\\
\{medical knowledge correctness evaluation criteria\}\\
\\
\#\# Below are two doctors' responses to a patient's question about a image. Now you need to analyze each response one by one then give a score between 0-5 to each response about each criterion above.\\
\\
\#\#\# Image And Question:\\
<image>\\
\{question\}\\
\\
\#\#\# Response 1:\\
\{response\_1\}\\
\\
\#\#\# Response 2:\\
\{response\_2\}\\
\\
\#\#\# Medical Knowledge Correctness Evaluation:\\
\{evaluation\}\\

\textbf{Patient Question Relevance:}\\
\#\#\# Instruction:\\
\#\# Evaluation Criterion: (higher score means better performance)\\
\{patient question relevance evaluation criteria\}\\
\\
\#\# Below are two doctors' responses to a patient's question about an image. Now you need to analyze each response one by one then give a score between 0-5 to each response about each criterion above.\\
\\
\#\#\# Image And Question:\\
<image>\\
\{question\}\\
\\
\#\#\# Response 1:\\
\{response\_1\}\\
\\
\#\#\# Response 2:\\
\{response\_2\}\\
\\
\#\#\# Patient Question Relevance Evaluation:\\
\{evaluation\}\\

\textbf{Conclusion:}\\
\#\#\# Instruction:\\
\#\# Below are two doctors' responses to a patient's question about a image followed by some sub-domain evaluations about the responses, analyze two responses comprehensively and give a final score between 0-5 to each response about each criterion above.\\
\\
\#\#\# Image And Question:\\
<image>\\
\{question\}\\
\\
\#\#\# Response 1:\\
\{response\_1\}\\
\\
\#\#\# Response 2:\\
\{response\_2\}\\
\\
\#\#\# Sub-domain Evaluations:\\
\#\# Sub-domain Patient Question Relevance:\\
\{patient question relevance evaluation result\}\\
\#\# Sub-domain Medical Knowledge Correctness:\\
\{medical knowledge correctness evaluation result\}\\
\#\# Sub-domain Expression:\\
\{expression evaluation result\}\\
\\
\#\#\# Evaluation:\\
\{evaluation\}
\end{tcolorbox}

\section{Example of Collected Evaluations}
\label{Example of Collected Evaluation}
\begin{tcolorbox}[colback=gray!20,  width=\textwidth, fontupper=\fontsize{8pt}{9pt}\selectfont, fontlower=\fontsize{8pt}{9pt}\selectfont, before=\vspace{0.01cm}, breakable=true]
\textbf{Image}: \includegraphics[width=0.4\textwidth]{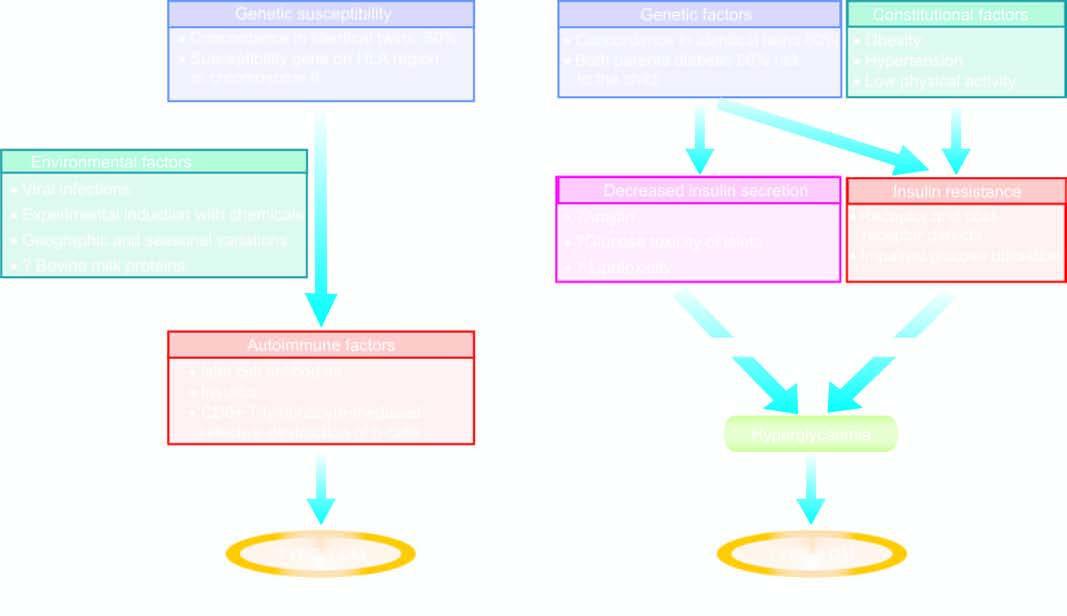}\\
\\
\textbf{Question}: what is involved in pathogenesis of two main types of diabetes mellitus?\\
\\
\textbf{Response 1}: The image likely illustrates the involvement of two main types of diabetes mellitus, type 1 and type 2, in the development of cardiovascular disease. It shows how the pathogenesis of these two types of diabetes can lead to complications such as coronary artery disease, heart failure, and stroke. The image may also highlight the role of various factors, such as hyperglycemia, inflammation, and advanced glycation end products, in the development and progression of these complications.\\
\\
\textbf{Response 2}: Type 1 diabetes is caused by an autoimmune reaction that destroys the insulin-producing beta cells in the pancreas. This leads to a lack of insulin production and the inability to regulate blood glucose levels. Type 2 diabetes is caused by a combination of factors, including insulin resistance, where the body does not respond properly to insulin, and beta cell dysfunction, where the pancreas does not produce enough insulin. This leads to elevated blood glucose levels.\\
\\
\textbf{Reference Answer}: chematic mechanisms.\\
\tcblower
\textbf{Expression Evaluation}: \\
Response 1:\\Criterion Clarity of Response:\\Analysis: The response provides a detailed explanation of the image, linking the pathogenesis of type 1 and type 2 diabetes to cardiovascular complications. It discusses various factors involved in the development of these complications.\\Score: 4\\Criterion Language Appropriateness:\\Analysis: The language used is suitable for patients, avoiding complex medical jargon and explaining terms clearly.\\Score: 4\\Criterion Tone and Empathy:\\Analysis: The response maintains a professional and informative tone, focusing on the medical aspects without conveying empathy towards the patient's emotional state.\\Score: 3\\Criterion Expression Integrity:\\Analysis: The response maintains coherence and consistency throughout, effectively connecting the pathogenesis of diabetes to cardiovascular disease.\\Score: 4\\\\Response 2:\\Criterion Clarity of Response:\\Analysis: The response clearly explains the pathogenesis of type 1 and type 2 diabetes, detailing the autoimmune destruction of beta cells in type 1 and the combination of factors in type 2.\\Score: 4\\Criterion Language Appropriateness:\\Analysis: The language is appropriate for patients, avoiding excessive medical jargon and explaining terms effectively.\\Score: 4\\Criterion Tone and Empathy:\\Analysis: The response lacks empathy towards the patient's emotional state, focusing solely on the medical information.\\Score: 2\\Criterion Expression Integrity:\\Analysis: The response maintains coherence and consistency, effectively explaining the mechanisms of both types of diabetes.\\Score: 4\\
\\
\textbf{Medical Knowledge Correctness Evaluation}: \\
Response 1:\\Criterion Factual Accuracy:\\Analysis: The response inaccurately describes the image as illustrating the involvement of type 1 and type 2 diabetes in cardiovascular disease, which is not directly related to the pathogenesis of diabetes mellitus.\\Score: 1\\Criterion Up-to-date Information:\\Analysis: The information provided is outdated and not in line with the current understanding of diabetes pathogenesis.\\Score: 0\\Criterion Handling Uncertainty:\\Analysis: The response does not acknowledge any uncertainty and presents misleading information.\\Score: 0\\\\Response 2:\\Criterion Factual Accuracy:\\Analysis: The response accurately describes the pathogenesis of type 1 and type 2 diabetes, including the autoimmune destruction of beta cells in type 1 and the combination of insulin resistance and beta cell dysfunction in type 2.\\Score: 5\\Criterion Up-to-date Information:\\Analysis: The information provided is current and reflects the latest medical knowledge on diabetes pathogenesis.\\Score: 5\\Criterion Handling Uncertainty:\\Analysis: The response confidently presents the information without any uncertainty.\\Score: 4\\
\\
\textbf{Patient Question Relevance Evaluation}: \\
Response 1:\\Criterion Context Awareness:\\Analysis: The response discusses the involvement of type 1 and type 2 diabetes in the development of cardiovascular disease, which is not directly related to the pathogenesis of diabetes mellitus.\\Score: 2\\Criterion Relevance to Patient's Condition:\\Analysis: The response does not specifically address the patient's condition or provide tailored information.\\Score: 1\\Criterion Addressing Multiple Concerns:\\Analysis: The response attempts to address the pathogenesis of both type 1 and type 2 diabetes but focuses more on cardiovascular complications.\\Score: 2\\\\Response 2:\\Criterion Context Awareness:\\Analysis: The response accurately describes the pathogenesis of type 1 and type 2 diabetes, focusing on the destruction of beta cells and insulin resistance.\\Score: 4\\Criterion Relevance to Patient's Condition:\\Analysis: The response provides relevant information about the pathogenesis of diabetes mellitus, tailored to the patient's question.\\Score: 4\\Criterion Addressing Multiple Concerns:\\Analysis: The response effectively addresses the pathogenesis of both type 1 and type 2 diabetes, covering multiple concerns comprehensively.\\Score: 4\\
\\
\textbf{Conclusion Evaluation}:\\
Analysis:\\Response 1: The response inaccurately describes the image and focuses more on cardiovascular complications rather than the pathogenesis of diabetes mellitus. It lacks relevance to the patient's question and does not provide tailored information. The information presented is outdated and misleading, scoring low in factual accuracy and up-to-date information.\\Response 2: In contrast, Response 2 accurately describes the pathogenesis of type 1 and type 2 diabetes, providing relevant and tailored information to the patient's question. It covers multiple concerns comprehensively and reflects the latest medical knowledge on diabetes pathogenesis. The response is clear, uses appropriate language, and maintains coherence and consistency.\\\\Final Score:\\Response 1: 1\\Response 2: 4
\end{tcolorbox}

\newpage
\section{Word Co-occurrence Graphs}
\label{Words Co-occurrence Graph}
\begin{figure}[htbp]
\centering
\begin{subfigure}{\linewidth}
    \centering
    \begin{subfigure}[b]{0.49\linewidth}
        
        \includegraphics[width=\linewidth]{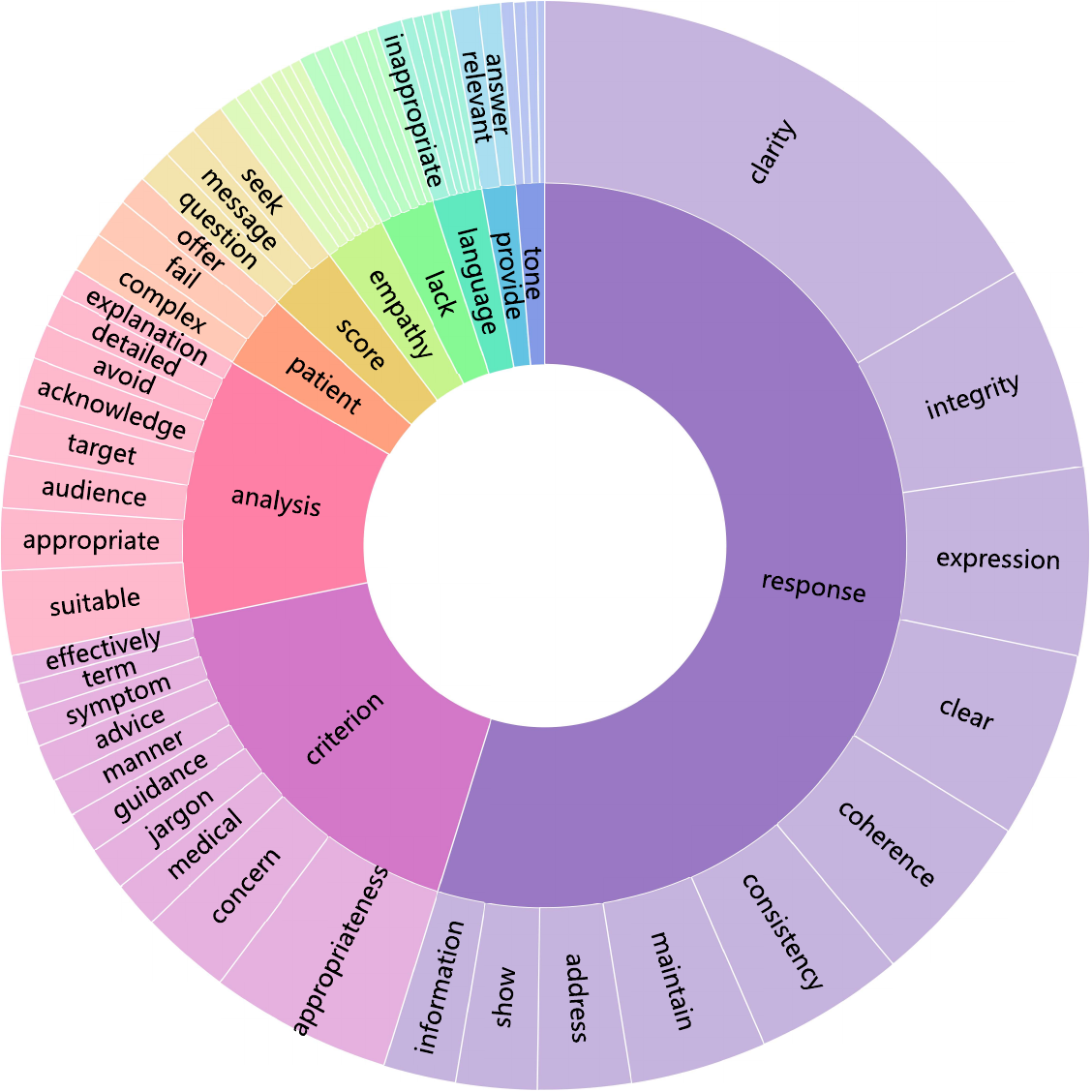}
    \end{subfigure}
    \hfill 
    \begin{subfigure}[b]{0.49\linewidth}
        \includegraphics[width=\linewidth]{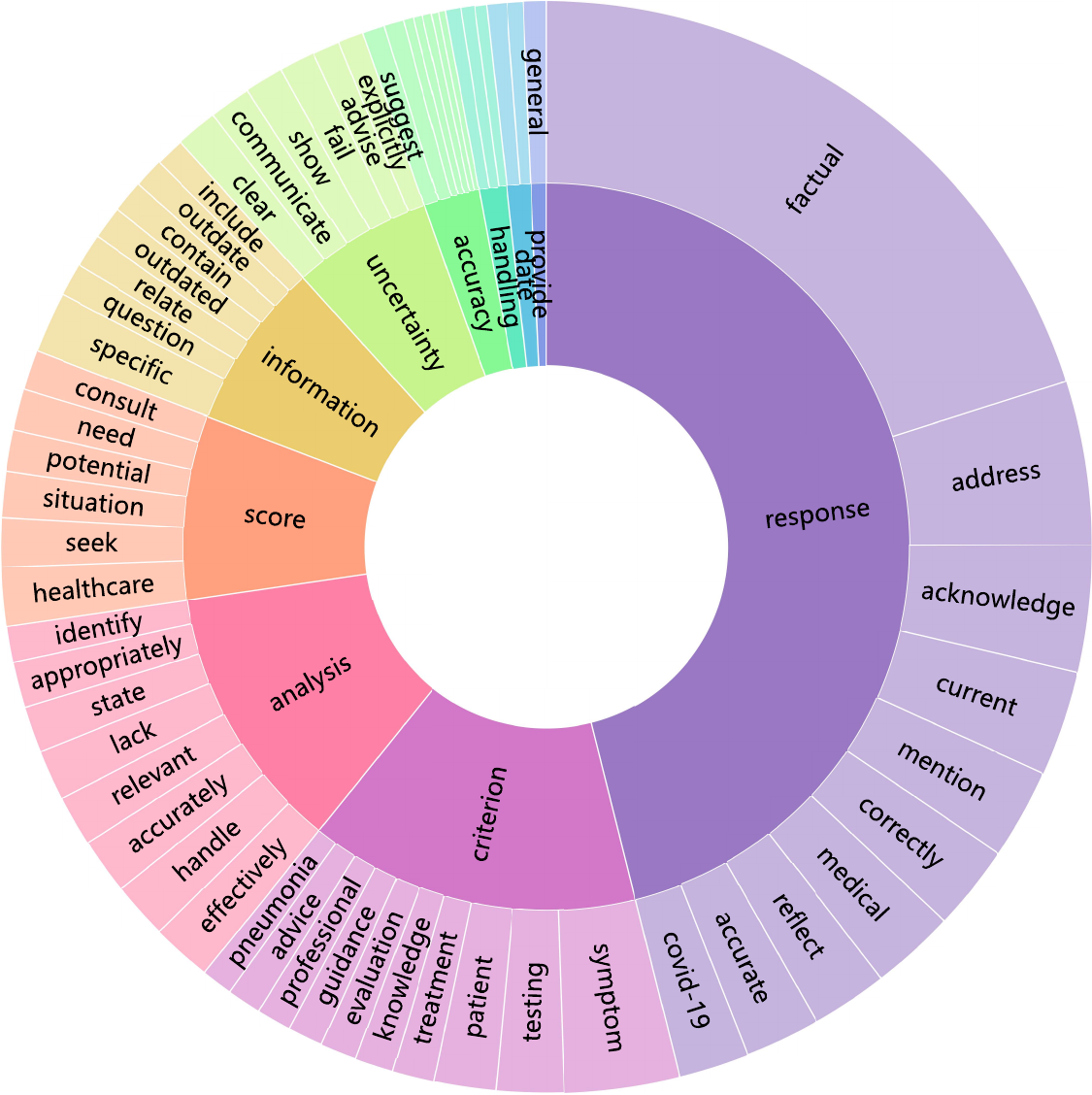}
    \end{subfigure}
\end{subfigure}

\begin{minipage}[b]{0.49\linewidth}
    \centering
    \text{Expression}
\end{minipage}
\hfill
\begin{minipage}[b]{0.49\linewidth}
    \centering
    \text{Medical Knowledge Correctness}
\end{minipage}

\vspace{10mm}

\begin{subfigure}{\linewidth}
    \begin{subfigure}[b]{0.49\linewidth}
        \centering
        \includegraphics[width=\linewidth]{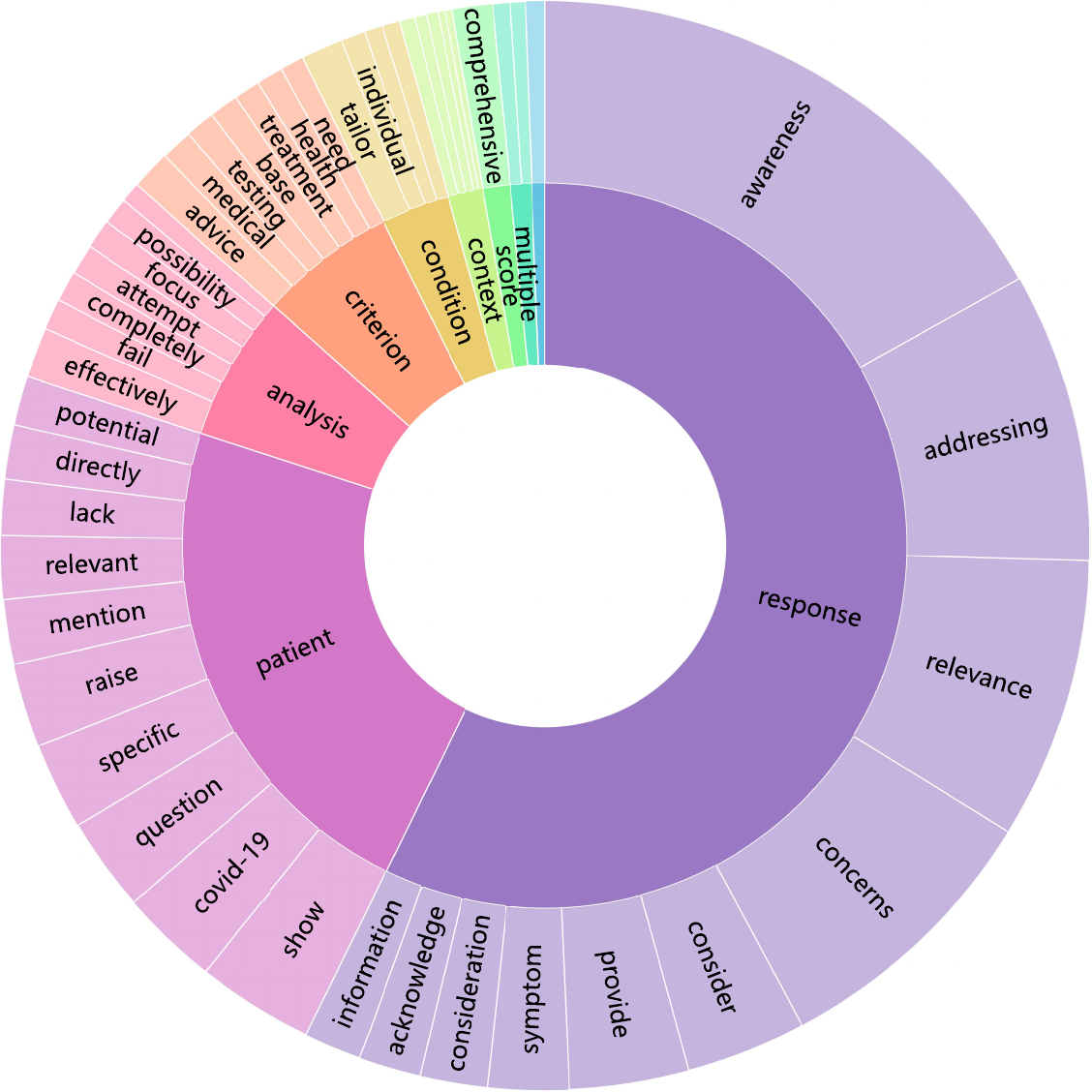}
    \end{subfigure}
    \hfill
    \begin{subfigure}[b]{0.49\linewidth}
        \centering
        \includegraphics[width=\linewidth]{Pictures/Data/conclusion.pdf}
    \end{subfigure}
\end{subfigure}

\begin{minipage}[b]{0.49\linewidth}
    \centering
    \text{Patient Question Relevance}
\end{minipage}
\hfill
\begin{minipage}[b]{0.49\linewidth}
    \centering
    \text{Conclusion}
\end{minipage}
\end{figure}

\newpage
\section{Results}
\subsection{Detailed Main Results}
\label{Detailed Main Results}
\begin{table}[H]
  \centering
  \resizebox{\linewidth}{!}{
    \begin{tabular}{llccccccccccc}
    \Xhline{1.0pt}
    \multicolumn{1}{l}{\multirow{2}[1]{*}{Model}} & \multirow{2}[1]{*}{Dataset} & \multicolumn{1}{c}{\multirow{2}[1]{*}{Conclusion}} & \multicolumn{4}{c}{EXP} & \multicolumn{3}{c}{MKC} & \multicolumn{3}{c}{PQR} \\
    \cmidrule(lr){4-7} \cmidrule(lr){8-10} \cmidrule(lr){11-13}
          & \multicolumn{1}{c}{} &       & \multicolumn{1}{c}{CR} & \multicolumn{1}{c}{LA} & \multicolumn{1}{c}{TE} & \multicolumn{1}{c}{EI} & \multicolumn{1}{c}{FA} & \multicolumn{1}{c}{UI} & \multicolumn{1}{c}{HU} & \multicolumn{1}{c}{CA} & \multicolumn{1}{c}{RPC} & \multicolumn{1}{c}{AMC} \\
    \hline
    \multicolumn{1}{l}{\multirow{7}[0]{*}{LLaVA-v1.6-Mistral-7b}} & MedDialogue-EN & 64.22  & 66.47  & 71.57  & 66.47  & 68.24  & 62.55  & 63.14  & 61.96  & 64.71  & 64.12  & 62.45  \\
          & MedText  & 60.10  & 55.39  & 61.27  & 49.02  & 53.04  & 61.57  & 56.76  & 59.41  & 58.24  & 56.37  & 54.71  \\
          & MedBench & 56.90  & 51.72  & 48.28  & 48.28  & 49.43  & 46.55  & 43.10  & 41.38  & 46.55  & 54.60  & 47.13  \\
          & MeadowWikidoc & 54.18  & 53.09  & 42.30  & 45.09  & 51.52  & 44.36  & 43.27  & 24.48  & 55.64  & 49.94  & 58.06  \\
          & Path-VQA  & 57.60  & 1.80  & 2.20  & 3.50  & 2.30  & 50.30  & 43.60  & 45.40  & 55.40  & 43.20  & 50.60  \\
          & SLAKE & 72.70  & 56.40  & 47.00  & 50.30  & 31.40  & 56.10  & 50.70  & 49.80  & 67.10  & 55.50  & 58.50  \\
          & VQA-RAD & 70.05  & 53.55  & 56.35  & 71.07  & 56.09  & 60.15  & 58.12  & 61.42  & 45.69  & 30.96  & 55.08  \\
    \hline
    \multicolumn{1}{l}{\multirow{7}[0]{*}{LLaVA-v1.6-Vicuna-7b}} & MedDialogue-EN & 38.33  & 65.10  & 62.16  & 62.16  & 63.53  & 58.63  & 59.71  & 55.69  & 63.82  & 60.59  & 60.69  \\
          & MedText  & 45.59  & 61.67  & 60.39  & 61.08  & 59.41  & 58.92  & 48.63  & 52.75  & 57.84  & 57.25  & 55.29  \\
          & MedBench & 4.02  & 48.85  & 46.55  & 48.85  & 50.57  & 39.08  & 31.61  & 33.33  & 35.06  & 19.54  & 17.82  \\
          & MeadowWikidoc & 11.27  & 43.03  & 31.39  & 38.06  & 38.06  & 41.21  & 41.58  & 25.45  & 40.85  & 45.45  & 43.27  \\
          & Path-VQA  & 0.40  & 20.00  & 22.80  & 17.80  & 17.10  & 23.40  & 22.10  & 25.80  & 24.10  & 18.20  & 21.90  \\
          & SLAKE & 0.70  & 8.40  & 6.10  & 9.10  & 3.20  & 5.40  & 6.00  & 4.40  & 10.50  & 8.40  & 11.70  \\
          & VQA-RAD & 0.25  & 9.64  & 9.39  & 7.87  & 7.87  & 18.27  & 20.05  & 19.29  & 15.23  & 14.47  & 19.04  \\
    \hline
    \multicolumn{1}{l}{\multirow{7}[0]{*}{LLaVA-v1.6-Vicuna-13b}} & MedDialogue-EN & 58.63  & 71.27  & 69.61  & 68.82  & 71.08  & 64.22  & 62.06  & 61.47  & 66.27  & 64.51  & 64.51  \\
          & MedText  & 44.22  & 66.57  & 69.02  & 48.92  & 65.10  & 62.25  & 59.22  & 42.84  & 65.59  & 61.47  & 59.61  \\
          & MedBench & 6.90  & 50.00  & 53.45  & 48.85  & 51.72  & 46.55  & 44.25  & 36.78  & 40.80  & 44.83  & 39.66  \\
          & MeadowWikidoc & 57.94  & 47.64  & 36.24  & 38.18  & 46.67  & 49.45  & 49.21  & 23.88  & 52.85  & 45.82  & 49.21  \\
          & Path-VQA  & 5.70  & 38.00  & 38.30  & 31.80  & 39.20  & 48.70  & 43.30  & 47.00  & 63.90  & 49.00  & 48.60  \\
          & SLAKE & 14.70  & 14.20  & 11.80  & 15.50  & 10.70  & 9.00  & 4.00  & 7.40  & 16.00  & 12.50  & 12.60  \\
          & VQA-RAD & 3.81  & 15.99  & 16.75  & 15.23  & 14.97  & 44.92  & 43.40  & 49.49  & 41.88  & 29.44  & 36.04  \\
    \hline
    \multicolumn{1}{l}{\multirow{4}[0]{*}{Gemini-1.0-Pro}} & MedDialogue-EN & 76.47  & 72.75  & 71.86  & 75.39  & 72.84  & 64.02  & 62.84  & 64.71  & 74.02  & 72.75  & 72.25  \\
          & MedText  & 69.12  & 70.10  & 66.47  & 53.14  & 68.73  & 56.08  & 56.37  & 55.88  & 70.00  & 70.29  & 67.94  \\
          & MedBench & 68.39  & 62.64  & 54.02  & 59.77  & 59.20  & 43.10  & 39.08  & 45.40  & 61.49  & 59.20  & 60.92  \\
          & MeadowWikidoc & 69.21  & 65.21  & 51.64  & 53.09  & 64.12  & 52.85  & 54.67  & 22.91  & 66.55  & 55.15  & 68.12  \\
    \hline
    \multicolumn{1}{l}{\multirow{3}[0]{*}{Gemini-1.0-Pro-Vision}} & Path-VQA  & 68.00  & 66.50  & 66.90  & 61.50  & 65.10  & 48.60  & 42.60  & 42.70  & 60.20  & 46.00  & 51.00  \\
          & SLAKE & 82.30  & 71.80  & 69.60  & 75.10  & 59.50  & 57.60  & 50.20  & 50.60  & 79.80  & 60.30  & 61.20  \\
          & VQA-RAD & 79.44  & 83.25  & 75.63  & 79.19  & 75.89  & 57.36  & 49.49  & 51.52  & 79.19  & 54.82  & 56.09  \\
    \hline
    \multicolumn{1}{l}{\multirow{7}[0]{*}{Gemini-1.5-Flash}} & MedDialogue-EN & 3.04  & 6.76  & 6.76  & 6.57  & 6.86  & 11.18  & 11.08  & 11.57  & 19.90  & 20.39  & 20.29  \\
          & MedText  & 3.04  & 16.57  & 15.88  & 16.37  & 16.67  & 15.20  & 13.92  & 13.82  & 5.69  & 5.69  & 5.78  \\
          & MedBench & 13.79  & 8.05  & 8.05  & 8.05  & 8.05  & 36.78  & 37.36  & 39.08  & 15.52  & 14.94  & 15.52  \\
          & MeadowWikidoc & 0.24  & 8.73  & 8.36  & 8.00  & 8.36  & 15.88  & 14.18  & 9.58  & 15.64  & 14.18  & 17.45  \\
          & Path-VQA  & 62.40  & 56.00  & 56.00  & 56.40  & 56.30  & 54.80  & 49.90  & 50.90  & 69.80  & 50.30  & 50.00  \\
          & SLAKE & 62.40  & 58.20  & 50.50  & 68.70  & 30.90  & 53.40  & 55.10  & 56.80  & 68.00  & 53.50  & 59.10  \\
          & VQA-RAD & 74.37  & 79.44  & 74.62  & 84.01  & 79.19  & 55.58  & 59.64  & 61.93  & 77.16  & 32.49  & 55.33  \\
    \hline
    \multicolumn{1}{l}{\multirow{7}[0]{*}{Gemini-1.5-Pro}} & MedDialogue-EN & 22.45  & 4.71  & 4.71  & 4.71  & 4.90  & 4.22  & 4.90  & 4.51  & 33.33  & 32.65  & 34.90  \\
          & MedText  & 22.84  & 2.75  & 2.55  & 2.45  & 2.65  & 5.10  & 4.90  & 4.61  & 16.67  & 16.57  & 16.47  \\
          & MedBench & 25.29  & 1.72  & 1.72  & 1.72  & 1.72  & 2.87  & 3.45  & 2.87  & 13.22  & 11.49  & 14.37  \\
          & MeadowWikidoc & 40.48  & 0.85  & 0.97  & 0.85  & 1.09  & 4.00  & 4.24  & 4.24  & 24.61  & 20.61  & 24.97  \\
          & Path-VQA  &  24.90  & 9.50  & 9.30  & 9.90  & 9.00  & 9.10  & 8.70  & 9.20  & 44.20  & 28.80  & 32.20  \\
          & SLAKE & 30.60  & 2.90  & 2.10  & 2.90  & 2.40  & 22.00  & 22.70  & 20.30  & 21.90  & 20.40  & 23.30  \\
          & VQA-RAD & 45.94  & 3.05  & 3.05  & 2.79  & 3.30  & 15.74  & 19.54  & 17.51  & 17.51  & 12.44  & 14.97  \\
    \hline
    \multicolumn{1}{l}{\multirow{7}[0]{*}{GPT-4-Turbo}} & MedDialogue-EN & 79.80  & 74.49  & 72.45  & 78.57  & 73.47  & 75.76  & 70.71  & 72.73  & 70.71  & 71.72  & 75.76  \\
          & MedText  & 74.23  & 59.60  & 76.77  & 74.75  & 69.70  & 67.68  & 72.73  & 66.67  & 70.71  & 73.74  & 70.71  \\
          & MedBench & 72.45  & 73.74  & 58.59  & 75.76  & 77.78  & 74.75  & 69.70  & 72.73  & 77.78  & 79.80  & 78.79  \\
          & MeadowWikidoc & 80.61  & 67.68  & 56.57  & 70.71  & 67.68  & 69.70  & 65.66  & 71.72  & 76.77  & 69.70  & 79.80  \\
          & Path-VQA  & 78.57  & 81.82  & 69.70  & 78.79  & 76.77  & 57.58  & 51.52  & 48.48  & 79.80  & 61.62  & 55.56  \\
          & SLAKE & 75.51  & 72.73  & 69.70  & 82.83  & 66.67  & 58.59  & 44.44  & 47.47  & 77.78  & 62.63  & 65.66  \\
          & VQA-RAD & 81.63  & 84.69  & 77.55  & 82.65  & 82.65  & 59.18  & 55.10  & 55.10  & 80.61  & 70.41  & 57.14  \\
    \hline
    \multicolumn{1}{l}{\multirow{4}[0]{*}{PandaLM}} & MedDialogue-EN & 75.20  & \multicolumn{10}{c}{\multirow{4}[0]{*}{/}} \\
          & MedText  & 70.98  & \multicolumn{10}{c}{} \\
          & MedBench & 70.11  & \multicolumn{10}{c}{} \\
          & MeadowWikidoc & 65.58  & \multicolumn{10}{c}{} \\
    \hline
    \multicolumn{1}{l}{\multirow{7}[0]{*}{Ours (ACE-$M^3$)}} & MedDialogue-EN & 76.67  & 73.53  & 71.76  & 76.57  & 72.45  & 66.18  & 66.18  & 69.12  & 71.57  & 72.45  & 75.10  \\
          & MedText  & 75.69  & 74.90  & 73.24  & 78.24  & 74.61  & 64.02  & 62.25  & 66.47  & 73.92  & 73.24  & 74.61  \\
          & MedBench & 71.26  & 58.62  & 44.83  & 58.05  & 56.90  & 53.45  & 52.30  & 54.60  & 69.54  & 68.97  & 69.54  \\
          & MeadowWikidoc & 64.00  & 60.00  & 45.82  & 56.12  & 58.06  & 46.06  & 48.97  & 49.09  & 61.33  & 55.76  & 67.27  \\
          & Path-VQA  & 82.10  & 80.80  & 76.50  & 79.50  & 77.70  & 69.70  & 64.30  & 63.40  & 84.20  & 72.00  & 64.50  \\
          & SLAKE & 83.20  & 79.60  & 71.00  & 77.00  & 78.20  & 65.00  & 63.30  & 61.40  & 83.90  & 62.60  & 69.40  \\
          & VQA-RAD & 82.99  & 81.47  & 73.60  & 83.50  & 78.43  & 65.99  & 64.72  & 64.21  & 81.22  & 73.10  & 68.02  \\
    \Xhline{1pt}
    \end{tabular}
    }
    \caption{Detailed main results.}
\end{table}

\newpage

\subsection{Score Distribution}
\label{score distribution}

\begin{table}[htbp]
  \centering
  \resizebox{0.7\linewidth}{!}{
    \begin{tabular}{lrrrrrrrrrrr}
    \Xhline{1.0pt}
    \multicolumn{1}{l}{\multirow{2}[1]{*}{Score}} & \multicolumn{1}{c}{\multirow{2}[1]{*}{Conclusion}} & \multicolumn{4}{c}{EXP} & \multicolumn{3}{c}{MKC} & \multicolumn{3}{c}{PQR} \\
    \cmidrule(lr){3-6} \cmidrule(lr){7-9} \cmidrule(lr){10-12}
          & & \multicolumn{1}{c}{CR} & \multicolumn{1}{c}{LA} & \multicolumn{1}{c}{TE} & \multicolumn{1}{c}{EI} & \multicolumn{1}{c}{FA} & \multicolumn{1}{c}{UI} & \multicolumn{1}{c}{HU} & \multicolumn{1}{c}{CA} & \multicolumn{1}{c}{RPC} & \multicolumn{1}{c}{AMC} \\
    \hline
    0 point     & 19.58  & 15.61  & 12.03  & 15.24  & 15.13  & 21.74  & 31.31  & 31.53  & 20.53  & 31.75  & 35.47  \\
    1 point     & 20.85  & 5.74  & 6.64  & 15.55  & 13.74  & 20.96  & 11.84  & 13.32  & 11.51  & 11.01  & 10.58  \\
    2 points    & 18.16  & 6.76  & 4.66  & 7.80  & 7.34  & 5.75  & 6.07  & 6.22  & 7.79  & 7.70  & 8.97  \\
    3 points    & 16.15  & 7.09  & 12.90  & 9.70  & 8.68  & 6.55  & 4.70  & 3.62  & 9.43  & 8.09  & 8.06  \\
    4 points    & 25.06  & 57.18  & 59.40  & 48.68  & 50.25  & 8.01  & 5.44  & 8.19  & 46.95  & 38.05  & 34.04  \\
    5 points    & 0.19  & 7.63  & 4.38  & 3.03  & 4.85  & 36.99  & 40.65  & 37.13  & 3.79  & 3.41  & 2.89  \\
    \hline
    \end{tabular}
    }
  \label{tab:addlabel}
  \caption{Distribution of scores generated by the model across various evaluation criteria.}
\end{table}%

\subsection{Detailed Ablation Results}
\label{Detailed Ablation Results}
\begin{table}[H]
  \centering
  \resizebox{0.98\linewidth}{!}{
    \begin{tabular}{llccccccccccc}
    \Xhline{1pt}
    \multirow{2}[3]{*}{Model} & \multirow{2}[3]{*}{Dataset} & \multirow{2}[3]{*}{Conclusion} & \multicolumn{4}{c}{EXP} & \multicolumn{3}{c}{MKC} & \multicolumn{3}{c}{PQR} \\
    \cmidrule(lr){4-7} \cmidrule(lr){8-10} \cmidrule(lr){11-13}
          &  &  & CR & LA & TE & EI & FA & UI & HU & CA & RPC & AMC \\
    \hline
     \multicolumn{1}{l}{\multirow{7}[0]{*}{Ours (ACE-$M^3$)}} & MedDialogue-EN & 76.67  & 73.53  & 71.76  & 76.57  & 72.45  & 66.18  & 66.18  & 69.12  & 71.57  & 72.45  & 75.10  \\
          & MedText  & 75.69  & 74.90  & 73.24  & 78.24  & 74.61  & 64.02  & 62.25  & 66.47  & 73.92  & 73.24  & 74.61  \\
          & MedBench & 71.26  & 58.62  & 44.83  & 58.05  & 56.90  & 53.45  & 52.30  & 54.60  & 69.54  & 68.97  & 69.54  \\
          & MedicalMeadowWikidoc & 64.00  & 60.00  & 45.82  & 56.12  & 58.06  & 46.06  & 48.97  & 49.09  & 61.33  & 55.76  & 67.27  \\
          & Path-VQA  & 82.10  & 80.80  & 76.50  & 79.50  & 77.70  & 69.70  & 64.30  & 63.40  & 84.20  & 72.00  & 64.50  \\
          & SLAKE & 83.20  & 79.60  & 71.00  & 77.00  & 78.20  & 65.00  & 63.30  & 61.40  & 83.90  & 62.60  & 69.40  \\
          & VQA-RAD & 82.99  & 81.47  & 73.60  & 83.50  & 78.43  & 65.99  & 64.72  & 64.21  & 81.22  & 73.10  & 68.02  \\
    \hline
     \multicolumn{1}{l}{\multirow{7}[0]{*}{w/o Reward Token}} & MedDialogue-EN & 76.67  & 73.73  & 72.16  & 78.24  & 74.71  & 66.47  & 66.37  & 66.96  & 71.08  & 71.96  & 74.02  \\
          & MedText  & 75.39  & 70.98  & 73.73  & 76.86  & 73.43  & 63.43  & 62.35  & 66.27  & 74.31  & 72.45  & 74.12  \\
          & MedBench & 67.82  & 58.62  & 48.85  & 57.47  & 55.17  & 43.10  & 38.51  & 47.13  & 64.37  & 66.67  & 64.94  \\
          & MedicalMeadowWikidoc & 64.61  & 59.88  & 44.85  & 58.18  & 58.79  & 47.27  & 49.45  & 48.61  & 59.64  & 53.70  & 67.39  \\
          & Path-VQA  & 80.30  & 78.80  & 78.10  & 79.80  & 76.90  & 68.50  & 64.30  & 64.50  & 80.90  & 69.40  & 62.40  \\
          & SLAKE & 80.90  & 78.60  & 67.60  & 76.20  & 76.90  & 66.10  & 62.80  & 62.70  & 81.70  & 61.30  & 69.80  \\
          & VQA-RAD & 82.99  & 80.71  & 70.56  & 80.96  & 77.66  & 66.75  & 66.75  & 65.99  & 75.89  & 69.29  & 63.20  \\
    \hline
    \multicolumn{1}{l}{\multirow{7}[0]{*}{w/o DPO}} & MedDialogue-EN & 75.69  & 73.63  & 72.25  & 75.78  & 72.84  & 66.57  & 66.67  & 68.92  & 70.20  & 70.98  & 73.14  \\
          & MedText  & 74.12  & 74.51  & 72.55  & 78.24  & 74.41  & 63.92  & 62.25  & 66.08  & 73.92  & 73.43  & 74.12  \\
          & MedBench & 66.67  & 58.62  & 45.98  & 59.20  & 56.32  & 46.55  & 40.80  & 47.70  & 69.54  & 67.24  & 68.97  \\
          & MedicalMeadowWikidoc & 62.91  & 60.12  & 45.82  & 56.61  & 56.97  & 44.85  & 48.48  & 47.76  & 57.45  & 50.91  & 63.64  \\
          & Path-VQA  & 82.70  & 80.60  & 75.90  & 78.80  & 77.80  & 69.20  & 64.90  & 63.80  & 53.60  & 43.60  & 40.20  \\
          & SLAKE & 82.60  & 79.40  & 69.80  & 77.00  & 77.40  & 66.50  & 61.80  & 61.30  & 60.60  & 43.30  & 50.60  \\
          & VQA-RAD & 82.23  & 82.49  & 73.60  & 83.76  & 79.19  & 64.97  & 63.96  & 62.94  & 53.30  & 46.19  & 42.39  \\
    \Xhline{1pt}
    \end{tabular}
  }
  \caption{Detailed ablation results of training strategies.}
  \vspace{-6mm}
\end{table}

\begin{table}[H]
  \centering
  \resizebox{0.98\linewidth}{!}{
    \begin{tabular}{llccccccccccc}
    \Xhline{1pt}
    \multicolumn{1}{l}{\multirow{2}[1]{*}{Frozen Layers}} & \multirow{2}[1]{*}{Dataset} & \multicolumn{1}{c}{\multirow{2}[1]{*}{Conclusion}} & \multicolumn{4}{c}{EXP} & \multicolumn{3}{c}{MKC} & \multicolumn{3}{c}{PQR} \\
    \cmidrule(lr){4-7} \cmidrule(lr){8-10} \cmidrule(lr){11-13}
          & \multicolumn{1}{c}{} &       & \multicolumn{1}{c}{CR} & \multicolumn{1}{c}{LA} & \multicolumn{1}{c}{TE} & \multicolumn{1}{c}{EI} & \multicolumn{1}{c}{FA} & \multicolumn{1}{c}{UI} & \multicolumn{1}{c}{HU} & \multicolumn{1}{c}{CA} & \multicolumn{1}{c}{RPC} & \multicolumn{1}{c}{AMC} \\
    \hline
    \multicolumn{1}{l}{\multirow{7}[0]{*}{Ours(Layers 0-24)}} & MedDialogue-EN & 76.67  & 73.53  & 71.76  & 76.57  & 72.45  & 66.18  & 66.18  & 69.12  & 71.57  & 72.45  & 75.10  \\
          & MedText  & 75.69  & 74.90  & 73.24  & 78.24  & 74.61  & 64.02  & 62.25  & 66.47  & 73.92  & 73.24  & 74.61  \\
          & MedBench & 71.26  & 58.62  & 44.83  & 58.05  & 56.90  & 53.45  & 52.30  & 54.60  & 69.54  & 68.97  & 69.54  \\
          & MedicalMeadowWikidoc & 64.00  & 60.00  & 45.82  & 56.12  & 58.06  & 46.06  & 48.97  & 49.09  & 61.33  & 55.76  & 67.27  \\
          & Path-VQA  & 82.10  & 80.80  & 76.50  & 79.50  & 77.70  & 69.70  & 64.30  & 63.40  & 84.20  & 72.00  & 64.50  \\
          & SLAKE & 83.20  & 79.60  & 71.00  & 77.00  & 78.20  & 65.00  & 63.30  & 61.40  & 83.90  & 62.60  & 69.40  \\
          & VQA-RAD & 82.99  & 81.47  & 73.60  & 83.50  & 78.43  & 65.99  & 64.72  & 64.21  & 81.22  & 73.10  & 68.02  \\
    \hline
    \multicolumn{1}{l}{\multirow{7}[0]{*}{Layers 0-15}} & MedDialogue-EN & 76.57  & 74.71  & 74.31  & 77.84  & 75.00  & 68.33  & 66.96  & 68.63  & 71.57  & 70.78  & 74.31  \\
          & MedText  & 75.69  & 77.16  & 73.63  & 78.14  & 76.67  & 64.71  & 63.14  & 65.69  & 75.69  & 74.61  & 74.51  \\
          & MedBench & 65.52  & 59.20  & 47.70  & 55.75  & 55.75  & 48.28  & 47.13  & 47.70  & 63.79  & 66.67  & 65.52  \\
          & MedicalMeadowWikidoc & 65.09  & 61.21  & 47.27  & 60.00  & 59.52  & 48.24  & 46.79  & 51.76  & 60.85  & 55.15  & 66.67  \\
          & Path-VQA  & 81.20  & 83.20  & 80.50  & 81.70  & 81.70  & 69.70  & 67.50  & 67.50  & 82.90  & 71.40  & 65.70  \\
          & SLAKE & 80.10  & 83.10  & 72.90  & 76.60  & 80.00  & 62.00  & 61.50  & 61.10  & 81.00  & 62.30  & 70.10  \\
          & VQA-RAD & 83.25  & 82.99  & 74.37  & 84.26  & 80.46  & 67.01  & 63.96  & 62.44  & 78.68  & 70.05  & 61.93  \\
    \hline
    \multicolumn{1}{l}{\multirow{7}[0]{*}{Layers 0-7}} & MedDialogue-EN & 78.33  & 75.88  & 72.25  & 77.06  & 73.24  & 69.51  & 66.08  & 67.65  & 69.80  & 70.39  & 72.75  \\
          & MedText  & 75.20  & 75.98  & 75.10  & 79.02  & 74.71  & 68.53  & 65.10  & 63.92  & 70.59  & 69.41  & 70.98  \\
          & MedBench & 65.52  & 68.39  & 50.57  & 59.77  & 65.52  & 46.55  & 46.55  & 40.23  & 62.64  & 59.77  & 67.24  \\
          & MedicalMeadowWikidoc & 67.76  & 64.61  & 45.70  & 61.58  & 63.15  & 51.15  & 51.64  & 48.36  & 63.76  & 56.73  & 69.94  \\
          & Path-VQA  & 84.80  & 84.40  & 81.90  & 82.10  & 82.40  & 68.90  & 62.60  & 63.10  & 63.80  & 57.00  & 52.70  \\
          & SLAKE & 84.10  & 84.00  & 74.50  & 77.30  & 79.80  & 67.80  & 65.00  & 63.90  & 74.60  & 57.80  & 60.80  \\
          & VQA-RAD & 84.52  & 82.23  & 76.14  & 84.77  & 79.70  & 67.77  & 63.20  & 64.47  & 65.99  & 55.08  & 54.31  \\
    \hline
    \multicolumn{1}{l}{\multirow{7}[0]{*}{No Freeze}} & MedDialogue-EN & 80.29  & 73.04  & 69.61  & 75.69  & 68.92  & 69.80  & 67.75  & 66.86  & 75.29  & 73.73  & 77.65  \\
          & MedText  & 77.06  & 76.08  & 74.31  & 79.12  & 73.14  & 68.73  & 64.90  & 62.55  & 73.43  & 72.65  & 72.94  \\
          & MedBench & 67.24  & 59.20  & 45.98  & 54.60  & 56.90  & 44.83  & 42.53  & 44.25  & 64.94  & 60.92  & 66.67  \\
          & MedicalMeadowWikidoc & 67.15  & 64.00  & 49.70  & 63.52  & 59.88  & 49.33  & 52.24  & 51.27  & 66.55  & 58.55  & 68.61  \\
          & Path-VQA  & 83.50  & 83.70  & 79.80  & 75.40  & 79.20  & 72.00  & 69.50  & 68.20  & 81.30  & 69.30  & 63.30  \\
          & SLAKE & 82.60  & 83.00  & 71.80  & 76.70  & 77.00  & 68.30  & 63.10  & 64.40  & 81.80  & 62.00  & 63.80  \\
          & VQA-RAD & 80.20  & 81.47  & 74.87  & 83.50  & 76.40  & 70.05  & 69.29  & 68.53  & 78.17  & 68.02  & 59.90  \\
    \Xhline{1.0pt}
    \end{tabular}
    }
    \caption{Detailed results with different numbers of frozen layers.}
    \vspace{-10mm}
\end{table}

\newpage

\section{Annotation}
\label{Annotation Details}
\begin{figure}[htbp]
    \centering
    \resizebox{0.98\linewidth}{!}{
    \includegraphics{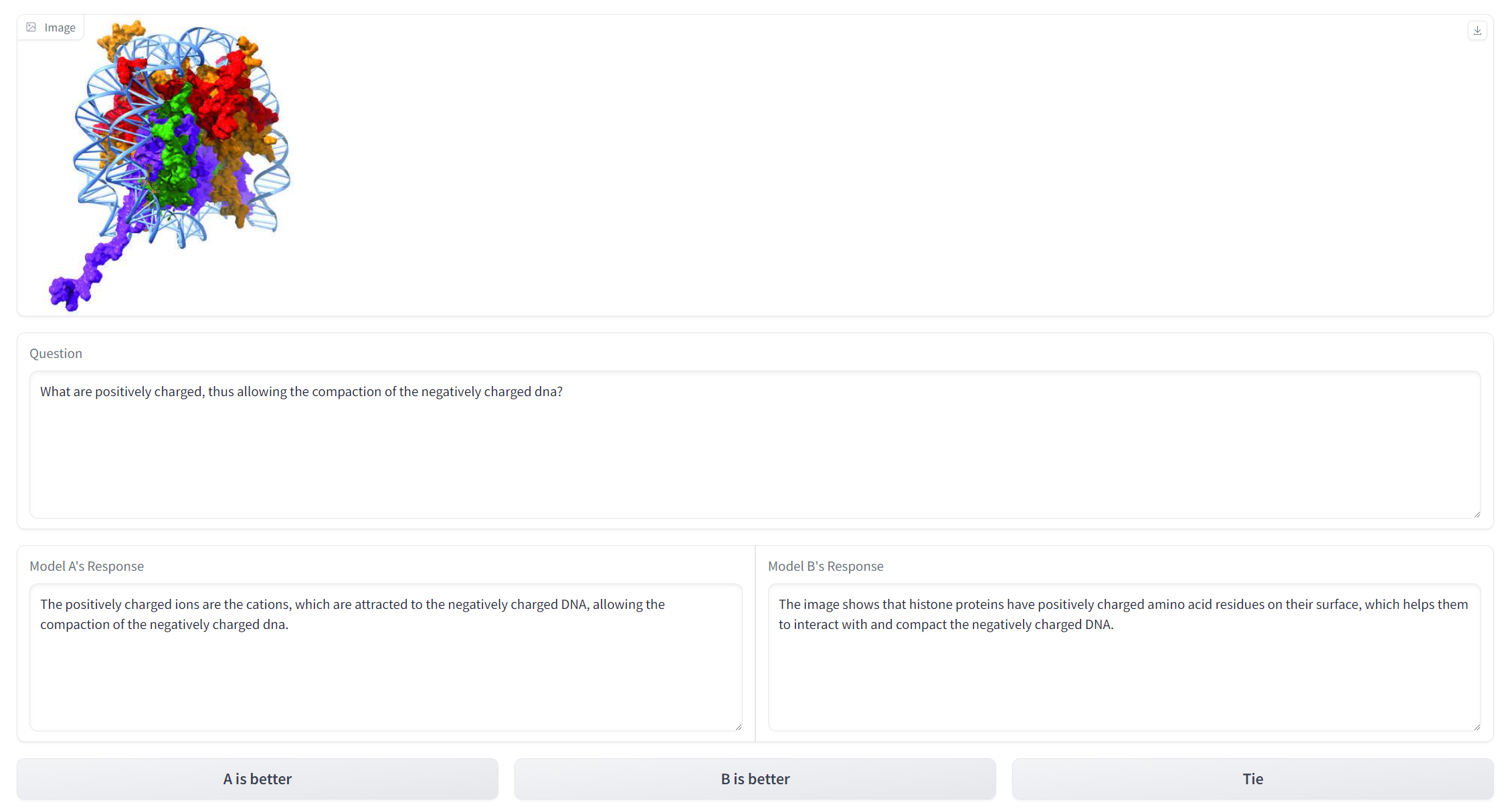}
    }
    \caption{The web-based user interface for conducting the human annotation.}
    \label{fig:annotation webui}
\end{figure}

Three professional annotators formed a committee, and for each case, the committee conducted a discussion lasting approximately 1-2 minutes based on the evaluation criteria provided in Section~\ref{Detailed Criteria} before making a final judgment. Upon completion, each annotator received a \$15 compensation per hour. The web interface used for annotation is shown in Figure~\ref{fig:annotation webui}.

\section{Further Evaluation Examples of ACE-$M^3$}
\label{more examples}
\begin{tcolorbox}[colback=gray!20,  width=\textwidth, fontupper=\fontsize{8pt}{9pt}\selectfont, fontlower=\fontsize{8pt}{9pt}\selectfont, breakable=true]
\textbf{Case \#1:}\\
\textbf{Image}: \includegraphics[width=0.6\textwidth]{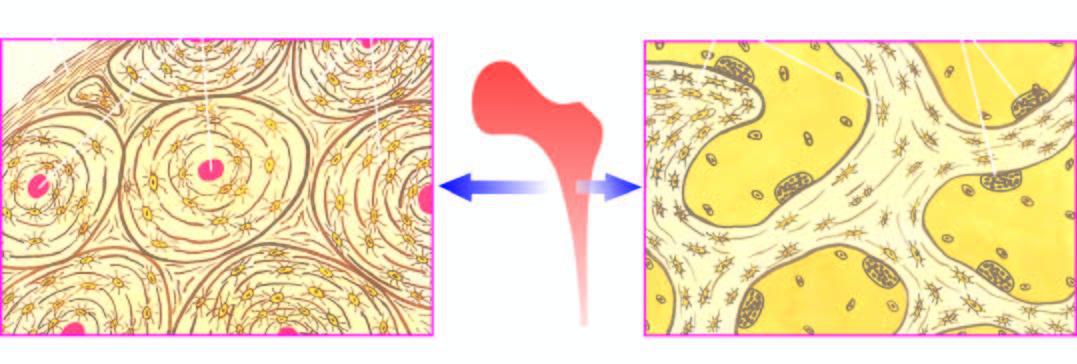}\\
\\
\textbf{Question}: does the cortical bone forming the outer shell show concentric lamellae along with osteocytic lacunae surrounding central blood vessels, while the trabecular bone forming the marrow space shows trabeculae with osteoclastic activity at the margins?\\
\\
\textbf{Response 1}: No\\
\\
\textbf{Response 2}: Yes, the cortical bone in the image forms a concentric lamellar structure with osteocytic lacunae surrounding central blood vessels.\\
\\
\textbf{Reference Answer} (not visible to \textbf{ACE-$M^3$}):  Yes\\
\tcblower
\textbf{ACE-$M^3$}:\\
\textbf{Expression Evaluation}:\\
Response 1:\\Criterion Clarity of Response:\\Analysis: The response is extremely brief and lacks any explanation or context. It does not provide any information to the patient.\\Score: 0\\Criterion Language Appropriateness:\\Analysis: The language used is appropriate, but the response is too simplistic and lacks detail.\\Score: 3\\Criterion Tone and Empathy:\\Analysis: The response lacks empathy as it does not address the patient's question or provide any helpful information.\\Score: 0\\Criterion Expression Integrity:\\Analysis: The response is inconsistent with the patient's question and does not maintain coherence or relevance.\\Score: 1\\Response 2:\\Criterion Clarity of Response:\\Analysis: The response is clear and provides a detailed explanation of the image, addressing the patient's question effectively.\\Score: 4\\Criterion Language Appropriateness:\\Analysis: The language used is suitable for the patient, explaining complex terms in a clear manner.\\Score: 4\\Criterion Tone and Empathy:\\Analysis: The response is empathetic and informative, showing understanding of the patient's query.\\Score: 4\\Criterion Expression Integrity:\\Analysis: The response is consistent and coherent, maintaining a logical flow of information.\\Score: 4\\
\\
\textbf{Medical Knowledge Correctness Evaluation}:\\
Response 1:\\Criterion Factual Accuracy:\\Analysis: The response is entirely incorrect as it states \"No\" when the correct answer is \"yes.\"\\Score: 0\\Criterion Up-to-date Information:\\Analysis: The response demonstrates up-to-date knowledge but is totally incorrect.\\Score: 0\\Criterion Handling Uncertainty:\\Analysis: The response confidently provides detailed information without any uncertainty.\\Score: 5\\Response 2:\\Criterion Factual Accuracy:\\Analysis: The response provides detailed and accurate information about the cortical and trabecular bone structures, correctly identifying the presence of concentric lamellae, osteocytic lacunae, and osteoclastic activity.\\Score: 5\\Criterion Up-to-date Information:\\Analysis: The response demonstrates up-to-date knowledge by accurately describing the features of cortical and trabecular bone.\\Score: 5\\Criterion Handling Uncertainty:\\Analysis: The response confidently provides detailed information without any uncertainty.\\Score: 5\\
\\
\textbf{Patient Question Relevance Evaluation}:\\
Response 1:\\Criterion Context Awareness:\\Analysis: The response lacks any relevant information or context to address the patient's question.\\Score: 0\\Criterion Relevance to Patient's Condition:\\Analysis: The response does not consider the patient's condition at all.\\Score: 0\\Criterion Addressing Multiple Concerns:\\Analysis: N/A\\Score: N/A\\Response 2:\\Criterion Context Awareness:\\Analysis: The response provides a detailed description of the image, mentioning the concentric lamellae and osteoclastic activity in the trabecular bone.\\Score: 4\\Criterion Relevance to Patient's Condition:\\Analysis: The response shows consideration for the patient's condition by addressing the specific features of the image.\\Score: 4\\Criterion Addressing Multiple Concerns:\\Analysis: The response effectively addresses multiple concerns by discussing both the cortical and trabecular bone in the image.\\Score: 4\\
\\
\textbf{Conclusion Evaluation}:\\
Analysis:\\Response 1: The response lacks relevance, accuracy, and clarity, providing no useful information to the patient.\\Response 2: The response is detailed, accurate, and clear, addressing the patient's question effectively.\\\\Final Score:\\Response 1: 0\\Response 2: 4
\end{tcolorbox}

\newpage
\begin{tcolorbox}[colback=gray!20,  width=\textwidth, fontupper=\fontsize{8pt}{9pt}\selectfont, fontlower=\fontsize{8pt}{9pt}\selectfont, before=\vspace{0.01cm}, breakable=true]
\textbf{Case \#2:}\\
\textbf{Image}: \includegraphics[width=0.2\textwidth]{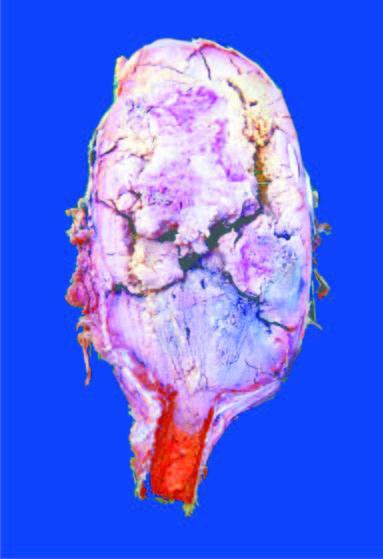}\\
\\
\textbf{Question}: What is expanded in the region of epiphysis?\\
\\
\textbf{Response 1}: The region of epiphysis is expanded.\\
\\
\textbf{Response 2}: The image shows the epiphysis expanded in the region of the femoral head. The epiphysis is the rounded, ball-like end of a long bone, and the femoral head is the upper part of the thigh bone (femur) that fits into the hip socket. The expansion in this region may be due to various reasons, such as a fracture, infection, or other pathological changes. Further evaluation and clinical correlation are needed to determine the cause and significance of this finding.\\
\\
\textbf{Reference Answer} (not visible to \textbf{ACE-$M^3$}):  end of the long bone.\\
\tcblower
\textbf{ACE-$M^3$}:\\
\textbf{Expression Evaluation}:\\
Response 1:\\Criterion Clarity of Response:\\Analysis: The response is extremely brief and lacks any explanation or context. It does not provide any information to the patient.\\Score: 0\\Criterion Language Appropriateness:\\Analysis: The language used is appropriate, but the response is too simplistic and lacks detail.\\Score: 3\\Criterion Tone and Empathy:\\Analysis: The response lacks empathy as it does not address the patient's concerns or provide any reassurance.\\Score: 0\\Criterion Expression Integrity:\\Analysis: The response is inconsistent with the patient's question and does not maintain coherence.\\Score: 1\\Response 2:\\Criterion Clarity of Response:\\Analysis: The response is clear and provides a detailed explanation of the image, addressing the patient's question effectively.\\Score: 4\\Criterion Language Appropriateness:\\Analysis: The language used is suitable for the patient, explaining medical terminologies in a clear manner.\\Score: 4\\Criterion Tone and Empathy:\\Analysis: The response shows empathy by acknowledging the patient's concerns and providing information in a reassuring manner.\\Score: 4\\Criterion Expression Integrity:\\Analysis: The response is consistent with the patient's question and maintains coherence throughout.\\Score: 5\\
\\
\textbf{Medical Knowledge Correctness Evaluation}:\\
Response 1:\\Criterion Factual Accuracy:\\Analysis: The response is entirely incorrect as it states that the region of epiphysis is expanded, which is not accurate.\\Score: 0\\Criterion Up-to-date Information:\\Analysis: The information provided is outdated and does not reflect the latest medical knowledge.\\Score: 0\\Criterion Handling Uncertainty:\\Analysis: The response does not acknowledge uncertainty or provide any indication of potential reasons for the expansion in the region of epiphysis.\\Score: 1\\Response 2:\\Criterion Factual Accuracy:\\Analysis: The response accurately identifies the region of epiphysis and provides additional information about its significance.\\Score: 4\\Criterion Up-to-date Information:\\Analysis: The response reflects up-to-date information about the region of epiphysis and its potential implications.\\Score: 5\\Criterion Handling Uncertainty:\\Analysis: The response effectively communicates the need for further evaluation and clinical correlation to determine the cause and significance of the expansion in the region of epiphysis.\\Score: 5\\
\\
\textbf{Patient Question Relevance Evaluation}:\\
Response 1:\\Criterion Context Awareness:\\Analysis: The response lacks any specific details or explanations related to the patient's question about the region of epiphysis.\\Score: 0\\Criterion Relevance to Patient's Condition:\\Analysis: The response does not consider the patient's condition or provide any relevant information.\\Score: 0\\Criterion Addressing Multiple Concerns:\\Analysis: The response does not address multiple concerns or provide comprehensive information.\\Score: 0\\Response 2:\\Criterion Context Awareness:\\Analysis: The response provides a detailed explanation of the region of epiphysis and its significance in the image.\\Score: 4\\Criterion Relevance to Patient's Condition:\\Analysis: The response considers the patient's condition by explaining the potential reasons for the expansion in the region of epiphysis.\\Score: 4\\Criterion Addressing Multiple Concerns:\\Analysis: The response addresses multiple concerns by discussing the significance of the expansion in the region of epiphysis and the need for further evaluation.\\Score: 4\\
\\
\textbf{Conclusion Evaluation}:\\
Analysis:\\Response 1: The response lacks accuracy, relevance, and clarity. It does not address the patient's question or provide any useful information. The lack of empathy and coherence further diminishes its effectiveness.\\Response 2: This response is detailed, accurate, and empathetic. It addresses the patient's concerns, provides relevant information, and maintains coherence throughout.\\\\Final Score:\\Response 1: 0\\Response 2: 4
\end{tcolorbox}

\end{document}